\documentclass{article} 

\newif\ifarxiv
\arxivtrue

\usepackage{iclr2027_conference,times}

\ifarxiv
    \iclrfinalcopy
\fi

\usepackage{amsmath,amsfonts,bm}

\def\eqref#1{equation~\ref{#1}}

\def\1{\bm{1}}

\DeclareMathAlphabet{\mathsfit}{\encodingdefault}{\sfdefault}{m}{sl}
\SetMathAlphabet{\mathsfit}{bold}{\encodingdefault}{\sfdefault}{bx}{n}

\usepackage{hyperref}
\usepackage{url}

\usepackage{graphicx}
\usepackage{booktabs}
\usepackage[table]{xcolor}
\usepackage[dvipsnames]{xcolor}  
\usepackage{pifont}
\usepackage{arydshln}
\usepackage{enumitem}
\usepackage{float}

\usepackage{algorithm}
\usepackage{algpseudocode}

\title{SliderEdit++: Improved Slider Trajectories in Instruction-Based Image Editing}

\author{Arman Zarei\textsuperscript{1,2}\thanks{Work done during internship at Adobe} , Sayan Nag\textsuperscript{1}, Bryan Wang\textsuperscript{1}, Franck Dernoncourt\textsuperscript{1}, Varun Manjunatha\textsuperscript{1}, \\ \textbf{Nedim Lipka\textsuperscript{1}}, \textbf{Soheil Feizi
\textsuperscript{2}}, \textbf{Samyadeep Basu\textsuperscript{1}}\\
  \textsuperscript{1}Adobe Research \hspace{5pt} \textsuperscript{2}University of Maryland}

\newcommand{\modelname}{SliderEdit++}
\newcommand{\benchmarkname}{SlideBench}

\newcommand{\cmark}{\ding{51}}
\newcommand{\xmark}{\ding{55}}

\begin{document}

\maketitle

\ifarxiv
    \pagestyle{plain}
\fi

\ifarxiv \vspace{-15pt} \fi
\begin{abstract}
\ifarxiv \vspace{-5pt} \fi

Slider-based control aims to give users continuous, fine-grained control over the intensity of individual edits in instruction-based image editing.
However, existing approaches fail to deliver reliable continuous controllability: 
traversing a slider can produce abrupt visual jumps and non-monotonic changes, while jointly controlled sliders can interfere with one another.
To systematically study these limitations, we introduce \emph{SlideBench}, a comprehensive benchmark for continuous image editing spanning diverse editing categories and equipped with a unified suite of metrics quantifying monotonic progression, perceptual continuity, and inter-slider disentanglement.
Building on these observations, we propose \modelname{}, which explicitly regularizes the intermediate editing trajectory rather than supervising only its endpoints. Our method combines velocity- and semantic-space trajectory regularization with spectral control to encourage smooth, predictable slider traversal, together with a disentanglement objective that reduces interference between independently controlled edits. On SlideBench, \modelname{} achieves state-of-the-art performance, improving the overall score by 18.5\% relative to the strongest baseline. At the same time, using a few-step editing backbone makes our method approximately $3\times$ to $20\times$ faster than existing slider-based approaches, enabling substantially more responsive interactive editing.
\ifarxiv
    \footnote[1]{Project page is available at: \href{https://armanzarei.github.io/SliderEditPlusPlus}{https://armanzarei.github.io/SliderEditPlusPlus}}
\fi

\end{abstract}
\ifarxiv \vspace{-10pt} \fi
\section{Introduction}
\label{sec:intro}
\ifarxiv \vspace{-5pt} \fi

Instruction-based image editing has advanced rapidly, with recent models such as FLUX-Kontext~\citep{labs2025flux}, Qwen-Image-Edit~\citep{wu2025qwen}, FLUX.2~\citep{flux-2-2025}, Nano Banana, and GPT-Image enable users to modify images through flexible natural-language instructions. 
While these models can strongly execute complex language instructions, they inherently produce a fixed, one-shot edit that often fails to precisely capture the user's nuanced visual intent.
Recent works therefore introduce \emph{continuous slider control} into instruction-based editing, allowing users to interactively vary the strength of an entire edit instruction or independently control individual components of a compositional instruction~\citep{zarei2026slideredit,parihar2026kontinuous,li2026veloedit,ekin2026unreasonable}.
Instead of committing to a fixed one-shot edit, users can traverse a continuum of possible outcomes and select the result that best matches their intent. Such continuous control is particularly appealing for interactive editing, where the desired appearance is often discovered through gradual adjustment rather than specified exactly beforehand.

\begin{figure}[t]
    \centering
    \begin{minipage}{0.345\textwidth}
        \centering
        \includegraphics[width=\linewidth]{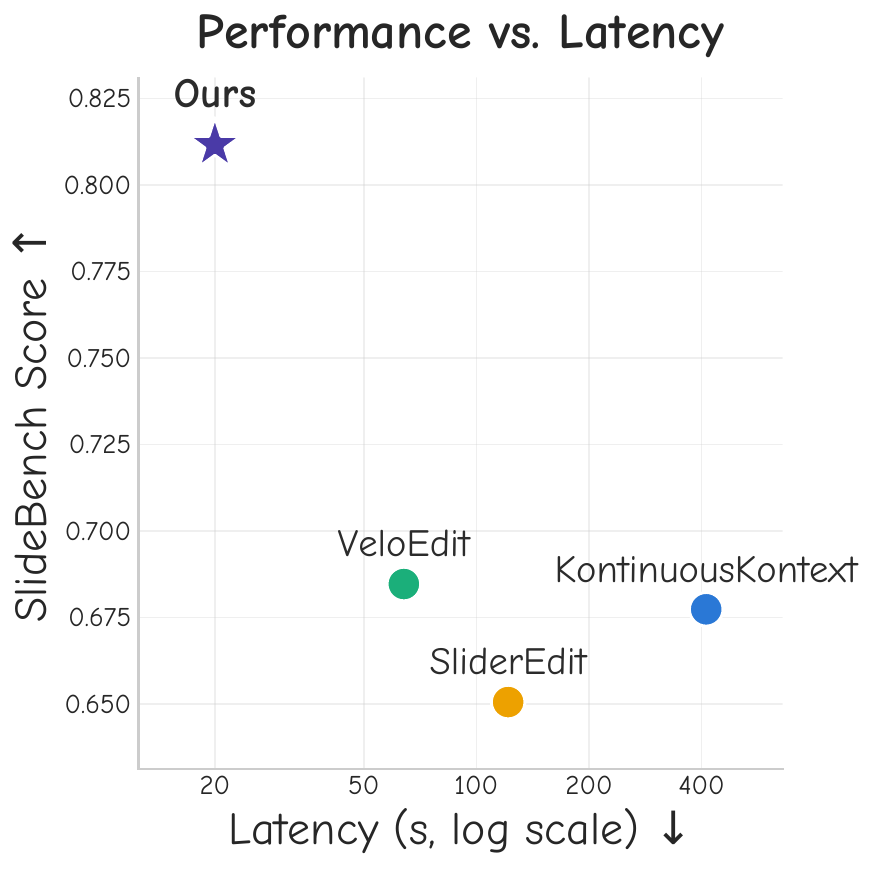}
        \vspace{-20pt}
        \caption{Our method achieves SOTA on \benchmarkname{} while being significantly faster.}
        \label{fig:performance_vs_speed_comparison}
    \end{minipage}
    \hfill
    \begin{minipage}{0.625\textwidth}
        \centering
        \includegraphics[width=\linewidth]{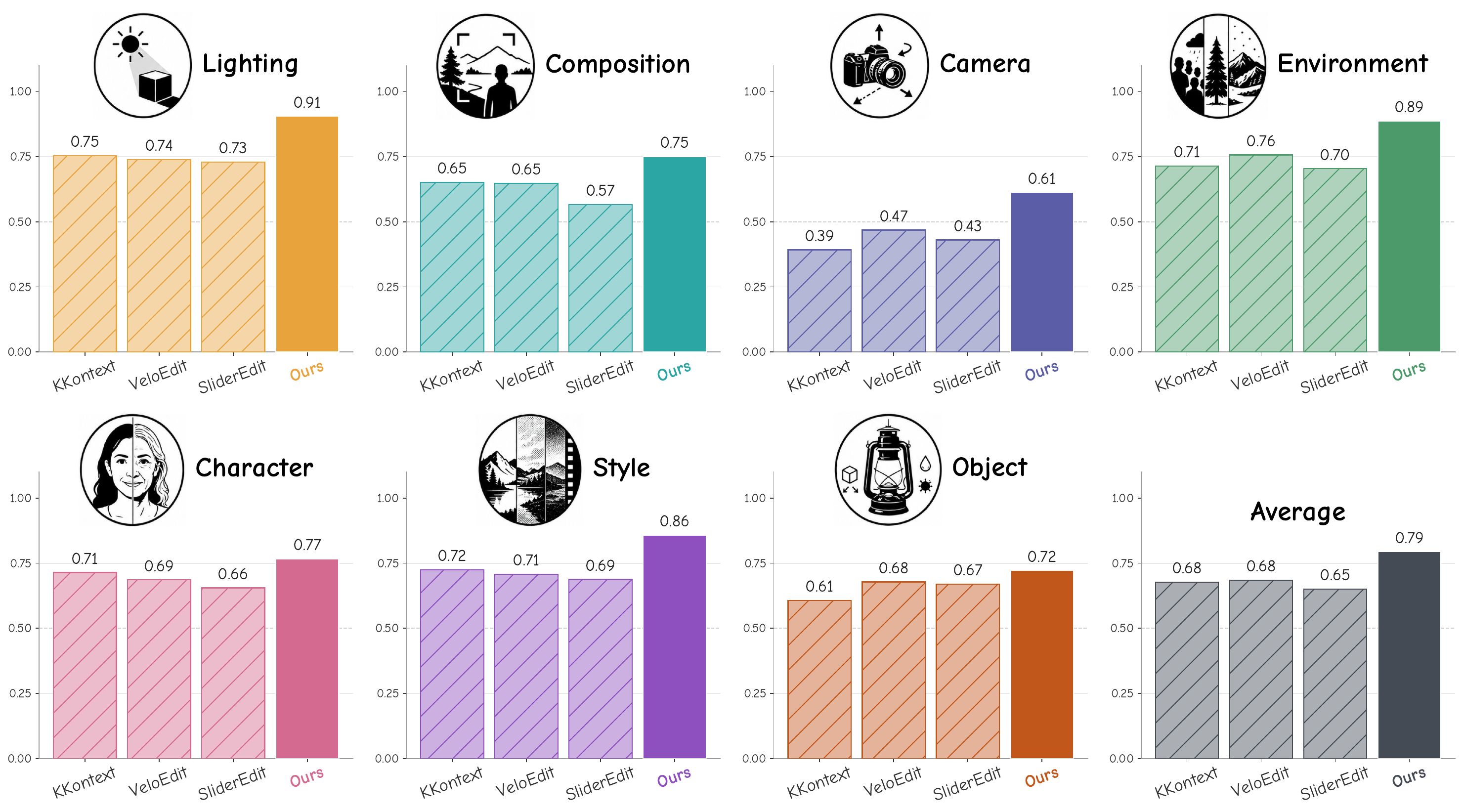}
        \vspace{-20pt}
        \caption{Comparison across \benchmarkname{} categories, spanning diverse continuous image editing scenarios, enabling a comprehensive evaluation of continuous controllability.}
        \label{fig:dataset_categories_and_models_performance}
    \end{minipage}
\ifarxiv \vspace{-5pt} \fi
\end{figure}

Despite the growing interest in slider-based image editing, their evaluation remains fragmented and inconsistent. Existing works typically rely on method-specific datasets, isolated subsets, and ad-hoc evaluation protocols, limiting meaningful comparison across approaches. Moreover, commonly used metrics often fail to capture the fine-grained perceptual behavior that defines effective continuous control. To address these limitations, we introduce \emph{SlideBench}, a comprehensive benchmark for continuous image editing spanning a diverse range of editing categories (Fig.~\ref{fig:dataset_categories_and_models_performance}) and fine-grained control axes that users may wish to adjust continuously in practice. Alongside the benchmark, we develop a unified and robust evaluation framework specifically designed to characterize slider behavior, with principled metrics for \emph{monotonic progression}, \emph{perceptual continuity}, and \emph{disentanglement}.

Evaluating existing methods on SlideBench reveals a consistent limitation: while performance varies across editing categories, slider trajectories are often far from truly continuous. Equal changes in slider value can induce highly uneven perceptual changes, including abrupt visual jumps (Fig.~\ref{fig:discontinuity_analysis}), while in the multi-slider setting, adjusting one control can unintentionally affect others. Motivated by these findings, we introduce \modelname{}, a training framework that explicitly regularizes slider trajectories while leveraging step-distilled editing backbones for efficient inference. Building on SliderEdit~\citep{zarei2026slideredit}, we introduce complementary trajectory regularizers that encourage smooth, uniform semantic progression across intermediate slider values, together with a disentanglement objective that encourages different sliders to act along orthogonal edit directions, preserving independence between controls. The resulting sliders are smooth, responsive, and independently controllable. On SlideBench, \modelname{} achieves state-of-the-art performance, improving the overall score by 18.5\% over the strongest baseline while being approximately 3$\times$ to 20$\times$ faster.

In summary, our main contributions are: \textbf{(i)} we introduce \emph{SlideBench}, a comprehensive benchmark for continuous instruction-based image editing spanning diverse editing categories, together with a unified evaluation suite for monotonic progression, perceptual continuity, and multi-slider disentanglement; \textbf{(ii)} we propose \emph{\modelname{}}, which explicitly regularizes slider trajectories during training to encourage smooth and uniformly progressing edits, while improving independence between multiple controls; and \textbf{(iii)} \modelname{} achieves \emph{SOTA performance on SlideBench}, improving the overall score by 18.5\% over the strongest baseline while operating approximately 3--20$\times$ faster.

\vspace{-5pt}
\section{Related Works}
\label{sec:related}
\vspace{-5pt}

Continuous control over semantic attributes has a long history in generative modeling. Early work on VAEs and GANs studied disentangled representations and semantically meaningful latent directions that enable continuous manipulation of visual attributes~\citep{higgins2017betavae,kim2018disentangling,shen2020interfacegan,harkonen2020ganspace,shen2021closed,abdal2021styleflow}. With text-to-image diffusion and rectified-flow models, this idea has evolved into explicit continuous controls over semantic attributes. Concept Sliders~\citep{gandikota2024concept} learn attribute-specific LoRA directions, while Prompt Sliders~\citep{sridhar2024prompt} and ~\citep{baumann2025continuous} manipulate directions in the text-embedding space. FluxSpace~\citep{dalva2024fluxspace} instead discovers semantically interpretable representations within rectified-flow transformers for fine-grained and disentangled semantic editing, while more recent works further automate the discovery of controllable directions or realize them through lightweight adapters~\citep{gandikota2025sliderspace,chiu2026text}.

More recently, continuous control has been extended to instruction-based image editing, where the goal is to modulate the strength of an arbitrary natural-language edit. SliderEdit~\citep{zarei2026slideredit} learns globally shared LoRA adapters by training them to suppress individual edit instructions, enabling both global and per-instruction sliders. Kontinuous Kontext~\citep{parihar2026kontinuous} instead explicitly conditions the editor on a scalar edit strength through a learned projector trained on synthesized strength-labeled data. Training-free approaches such as VeloEdit~\citep{li2026veloedit} and FlowSlider~\citep{endo2026flowslider} obtain continuous control by intervening directly in the model's velocity field, while \citep{ekin2026unreasonable} achieve steering through interpolation in the text-embedding space. More generic inference-time interventions such as SDEdit~\citep{meng2021sdedit} can also trade off source preservation and edit strength without dedicated slider training. Building on these advances, our method targets the limitations of existing sliders, improving their smoothness, consistency, and independence across controls.

\vspace{-3pt}
\section{\benchmarkname{} Evaluation Benchmark}
\label{sec:slidebench_evaluation}
\vspace{-3pt}
\ifarxiv \vspace{-5pt} \fi

In this section, we introduce \emph{SlideBench}, a comprehensive evaluation benchmark for continuous control in image editing. SlideBench is designed to assess a broad range of properties that characterize effective continuous editing, together with a robust suite of evaluation metrics that capture the fine-grained nuances of this setting.

\vspace{-5pt}
\subsection{Dataset Description}
\label{sec:dataset}
\vspace{-5pt}

To systematically evaluate continuous image editing, we construct a benchmark spanning a broad range of attributes that users may wish to adjust continuously in practice. The dataset is organized into seven high-level categories---\emph{Lighting}, \emph{Composition}, \emph{Camera}, \emph{Environment}, \emph{Character}, \emph{Style}, and \emph{Object}---covering $57$ fine-grained subcategories, with five samples per subcategory.
Each sample pairs a source image with an instruction that defines a continuous transformation between two semantic endpoints, with the same editing axis instantiated across diverse visual contexts to test generalization. 
The benchmark covers naturally gradual attributes, such as lighting and camera changes; edits that progressively introduce or strengthen visual content, such as crowd density or aging; and fine-grained character edits over facial attributes, where precise control over small visual changes is particularly important. Together, these categories provide broad coverage of continuous editing behavior and enable systematic comparison across methods.
Figure~\ref{fig:dataset_categories_and_models_performance} showcases the benchmark categories and compares the performance of baselines. For Additional details, refer to  Appendix~\ref{sec:app:eval_dataset}.

\vspace{-4pt}
\ifarxiv \vspace{-3pt} \fi
\subsection{Evaluation Metrics}
\label{sec:evaluation_metrics}
\ifarxiv \vspace{-2pt} \fi
\vspace{-4pt}

We evaluate each slider along two complementary dimensions: \emph{monotonic progression}, which measures whether the strength of the edit increases consistently as the slider advances, and \emph{continuity}, which measures whether these changes occur smoothly without abrupt transitions. For multi-slider models, we additionally evaluate \emph{disentanglement}, measuring whether varying one slider preserves the remaining attributes. We describe all the metrics below: 

\textbf{Monotonic Progression.}
Let $I_1,\dots,I_T$ denote the frames of a slider ordered by increasing edit strength. For each consecutive pair $(I_t, I_{t+1})$, we ask a set of powerful VLM judges which image applies the target edit more strongly. The two images are presented in randomized order to eliminate positional bias, and each comparison is scored as $1$ if the later frame is judged stronger and $0$ otherwise (including ties). The monotonic progression score is then the average accuracy over all consecutive pairs and all judges,
\vspace{-6pt}
\begin{equation}
\mathrm{MP}=\frac{1}{|\mathcal{J}|(T-1)}\sum{J\in\mathcal{J}}\sum_{t=1}^{T-1} m_t^J \in [0,1],
\qquad
m_t^J =
\begin{cases}
1, & \text{if } \hat{y}_t^J=t+1,\\
0, & \text{otherwise}
\end{cases}
\vspace{-3pt}
\end{equation}
where higher values indicate more consistently increasing edit strength along the slider.
We adopt pairwise VLM judgments because they capture fine-grained perceptual differences in edit strength that image--text similarity scores often fail to reflect in the general case. 

\textbf{Continuity.}
Let $d_t^\varphi$ denote the distance between consecutive frames $I_t$ and $I_{t+1}$ under the feature representation induced by a feature extractor $\varphi$ (e.g., DINOv2). We define the per-slider continuity score as:
$U^\varphi=\exp\!\left(-\operatorname{std}\!\left(\{d_t^\varphi\}_{t=1}^{T-1}\right)\right)\in(0,1]$,
where $U^\varphi=1$ indicates perfectly uniform transitions.
Since continuity alone cannot distinguish a smooth edit from a no-op slider (i.e., one that produces little or no visible change), we weight each sample by its overall edit strength $e\in[0,1]$, which measures how strongly the final frame exhibits the intended edit relative to the source image, as judged by a powerful VLM:
$\bar U^\varphi=\frac{1}{M}\sum_{i=1}^{M} U_i^\varphi\,e_i$.
The final continuity score $\bar U$ averages over all feature representations. 



\textbf{Multi-slider Disentanglement.}
We evaluate whether changing one slider preserves the remaining attributes. For each slider $k$, we fix all other sliders to obtain a 1D slice, and measure the alignment of every image in the slice with the captions corresponding to each non-varying attribute $j\neq k$. Disentanglement is defined as the mean absolute deviation of these alignment scores along the slice:
$D_{k\rightarrow j}=\frac{1}{A}\sum_{a=1}^{A}\left|s_j(I_a^{(k)})-\bar{s}_j^{(k)}\right|$,
where $s_j(\cdot)$ is the alignment score for attribute $j$ and $\bar{s}_j^{(k)}$ is its mean over the slice. We average $D_{k\rightarrow j}$ over all slices and slider pairs $(k,j)$, where lower values indicate better disentanglement. 

\begin{figure}[t]
    \centering
    \begin{minipage}{0.6\textwidth}
        \centering
        \includegraphics[width=\linewidth]{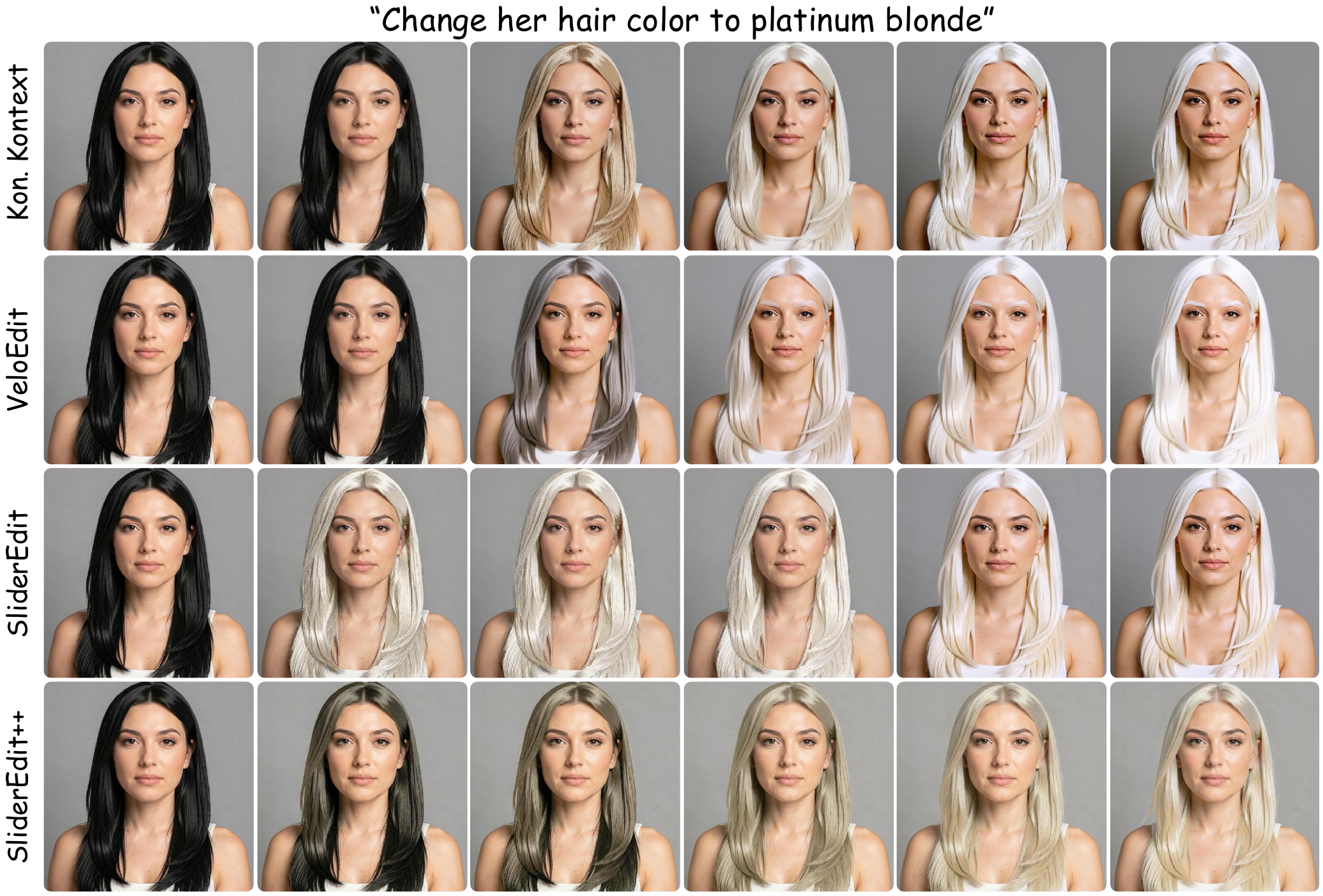}
    \end{minipage}
    \hfill
    \begin{minipage}{0.38\textwidth}
        \centering
        \includegraphics[width=\linewidth]{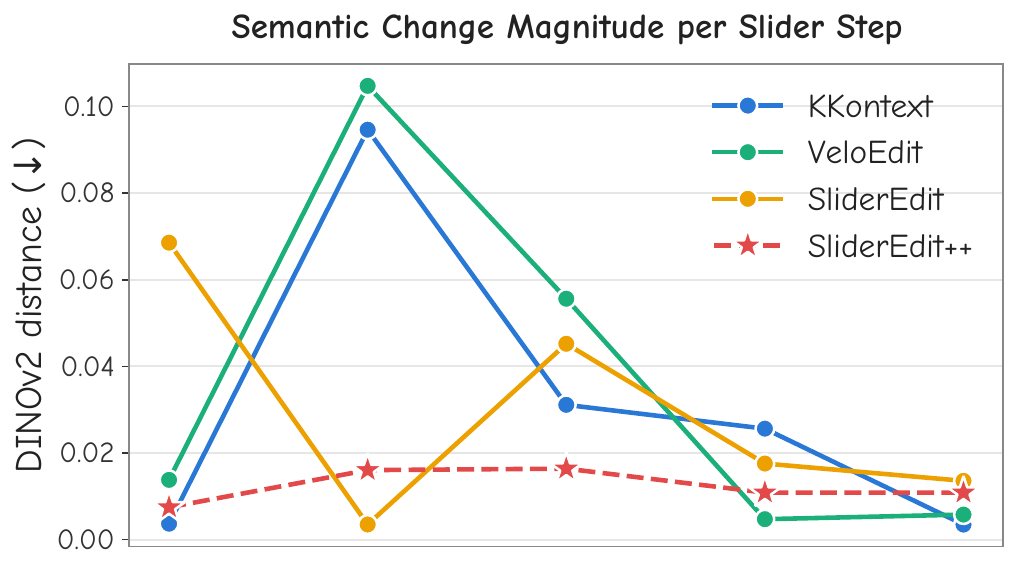}
        \includegraphics[width=\linewidth]{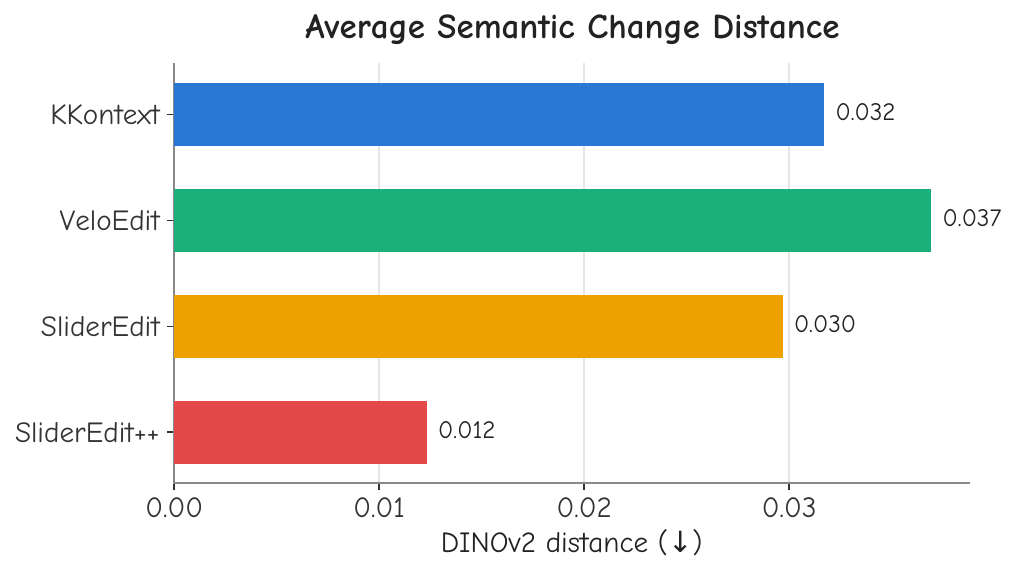}
    \end{minipage}
    \vspace{-10pt}
    \caption{\textbf{Slider Discontinuity Analysis}. Baseline image-editing sliders often exhibit abrupt transitions during interaction. The semantic change distance on the right quantifies this instability. In contrast, \modelname{} produces smoother sliders by addressing the issue during training.}
\label{fig:discontinuity_analysis}
\end{figure}

\section{Method: Improving Continuity and Disentanglement in Sliders}

Although recent image editing sliders provide intuitive and seemingly continuous control over semantic attributes, we observe that their editing trajectories are often far from smooth. As illustrated in Figure~\ref{fig:discontinuity_analysis}, traversing a slider frequently produces abrupt semantic transitions, where a small change in the slider value unexpectedly results in a disproportionately large change in the generated image (e.g., the sudden hair color transition in each method). Such discontinuities break the notion of continuous control, making fine-grained editing unstable and unpredictable. This phenomenon is not limited to qualitative observations. We quantify the semantic change between consecutive slider positions using the DINOv2 feature distance. Ideally, a well-behaved editing slider should induce \emph{uniform semantic progress}, where consecutive slider steps produce similar semantic changes throughout the entire trajectory. Instead, existing methods exhibit large fluctuations in semantic distance, revealing irregular editing dynamics (Figure~\ref{fig:discontinuity_analysis}, top right). These results suggest that current slider-based editors learn an inconsistent control space, where equal slider increments do not correspond to equal semantic changes.

Motivated by these observations, we propose \modelname{}, which explicitly regularizes the editing trajectory during training to encourage smooth and consistent semantic progression. As a result, our method produces stable, predictable slider behavior, yielding visually smooth transitions and substantially more uniform semantic changes across the entire editing trajectory. In the following sections, we describe the key components of our approach and show how they collectively enable the learning of a smooth and consistent editing space.


\vspace{-4pt}
\subsection{Preliminaries: Continuous Edit Slider using SliderEdit}
\label{sec:prelim}
\vspace{-4pt}

We build on SliderEdit~\citep{zarei2026slideredit}, which learns continuous editing controls by training adapters to progressively suppress the effect of an editing instruction.
Specifically, let $\epsilon(Z_t, X, P)$ denote the frozen base editing model's velocity prediction for noisy latent $Z_t$ at timestep $t$, given the source image latent $X$ and a text instruction $P$. SliderEdit injects a low-rank adapter at every layer $\ell$, $\Delta W^\ell = B^\ell A^\ell$, and defines a \emph{scaled} adapted model $M_\theta^{\alpha}$ that applies $\alpha\,\Delta W^\ell$ at each layer, with $\alpha\in[0,1]$ acting as the slider value: $\alpha=0$ leaves the base model's instruction-following edit untouched, while $\alpha=1$ is trained to fully \emph{suppress} it. Concretely, the adapter is trained with a Partial Prompt Suppression loss,
\vspace{-6pt}
\begin{equation}
\mathcal{L}_{\text{PPS}} = \mathbb{E}_{Z_t,t}\Big\| \epsilon_{M_\theta^{1}}(Z_t, X, P) \;-\; \epsilon(Z_t, X, P_\emptyset) \Big\|^2 ,
\label{eq:pps}
\vspace{-3pt}
\end{equation}
where $P_\emptyset$ is a neutral, edit-free prompt (e.g.\ ``keep the image the same''), and the right-hand side is a stop-gradient target computed with the base model alone. In words: at full strength, the adapter is pulled toward reproducing what the \emph{unedited} model would have predicted, so that sweeping $\alpha$ from $0\to1$ at inference sweeps the output continuously from ``fully edited'' to ``neutral''. In the multi-instruction setting, the same idea is applied \emph{per instruction} rather than to the whole prompt.

This objective supervises only the endpoints $\alpha=0$ and $\alpha=1$. Intermediate controls arise simply by continuously scaling the learned LoRA weights, without any direct supervision. Consequently, as shown in Figure~\ref{fig:discontinuity_analysis}, the resulting trajectories can exhibit discontinuities, which we address with the regularizers introduced below.

\begin{figure}[t]
    \centering
    \includegraphics[width=\textwidth]{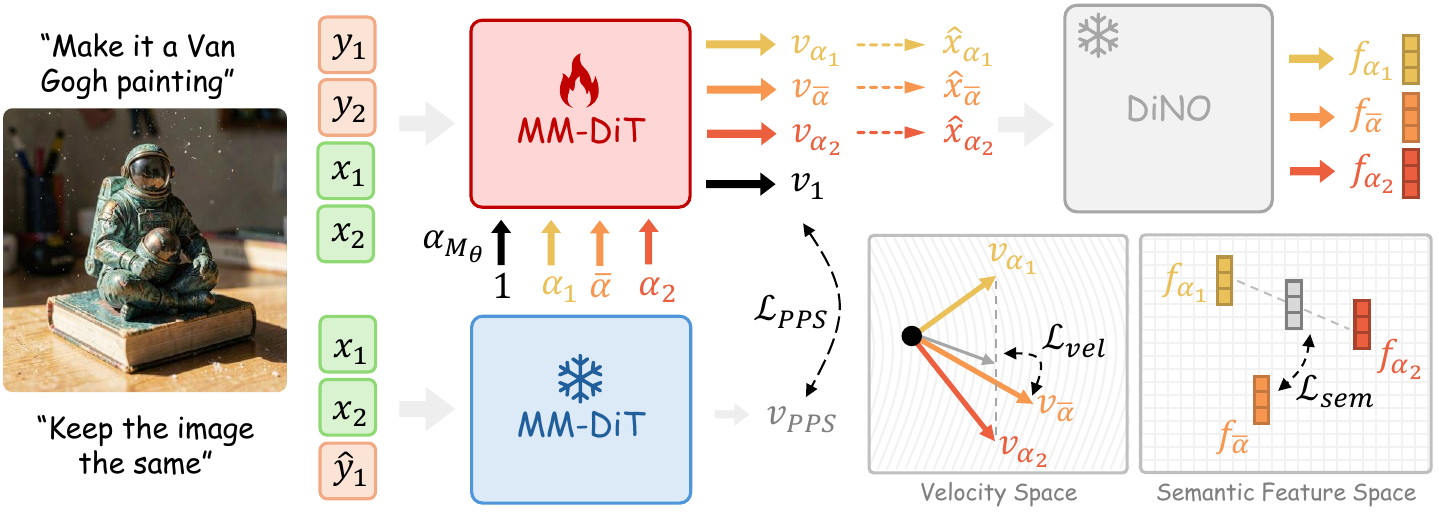}
    \vspace{-20pt}
    \caption{Overall pipeline of \modelname{} for improving the continuity in image editing sliders.}
    \label{fig:method}
\end{figure}

\vspace{-4pt}
\subsection{Towards Solving Continuity Issues in Sliders}
\label{sec:continuity}
\vspace{-4pt}

Because $\mathcal{L}_{\text{PPS}}$ only anchors the discrete endpoints $\alpha \in \{0,1\}$, the intermediate trajectory $\alpha \mapsto \epsilon_{M_\theta^\alpha}$ remains unconstrained. Consequently, two adapters with identical endpoint behavior can deviate arbitrarily at intermediate values, producing abrupt perceptual jumps rather than a smooth progression. To enforce uniform, monotonic behavior across $\alpha \in [0,1]$, we introduce three complementary regularizers, each targeting a different reason the interior of the slider can misbehave.

\textbf{Velocity-space midpoint regularizer.} SliderEdit observes that \emph{explicit} classifier-free guidance~\citep{ho2022classifier} -- linearly blending two model outputs, $\epsilon = (1-\alpha)\,\epsilon_{\text{edit}} + \alpha\,\epsilon_{\text{neutral}}$ -- gives highly smooth transitions almost by construction, since it is a literal convex combination of two vectors. Scaling LoRA \emph{weights} by $\alpha$ is not the same thing: $\epsilon_{M_\theta^\alpha}$ is a highly nonlinear function of $\alpha$ once it passes through many transformer layers, so linear interpolation of the \emph{weights} gives no guarantee of linear interpolation of the \emph{output}. We inject that missing guarantee directly as a training signal. For two random scales $\alpha_1,\alpha_2\sim\mathcal{U}[0,1]$ and their midpoint $\alpha_m=\tfrac{1}{2}(\alpha_1+\alpha_2)$, we penalize
\vspace{-5pt}
\begin{equation}
\mathcal{L}_{\text{vel}} = \mathbb{E}\Big\| \epsilon_{M_\theta^{\alpha_m}}(Z_t,X,P) - \tfrac{1}{2}\big(\operatorname{sg}[\epsilon_{M_\theta^{\alpha_1}}(Z_t,X,P)] + \operatorname{sg}[\epsilon_{M_\theta^{\alpha_2}}(Z_t,X,P)]\big)\Big\|^2,
\label{eq:vel}
\vspace{-3pt}
\end{equation}
where $\operatorname{sg}[\cdot]$ denotes the stop-gradient operator.
So the loss only shapes the \emph{interior} of the curve and leaves the endpoints -- already anchored by $\mathcal{L}_{\text{PPS}}$ -- untouched. Enforcing the midpoint's prediction to equal the average of its neighbors' predictions, for randomly sampled neighbors everywhere in $[0,1]$, is the discrete analogue of asking $\alpha\mapsto\epsilon_{M_\theta^\alpha}$ to be affine: exactly the property a trustworthy slider needs, equal steps in $\alpha$ should produce equal, predictable steps in the output.

\textbf{Semantic feature-space regularizer.} Smoothness in raw velocity space does not imply smoothness in what the user actually perceives. The map from latent velocity, through decoding, to a visible attribute (e.g.\ ``how much is the person smiling'') is highly nonlinear, and a small residual in $\epsilon$ can be perceptually invisible in one region of latent space and produce a visible jump in another. Since the slider is judged by what it looks like, not by its latent representation, we apply the same midpoint-consistency idea directly in a semantic feature space. Let $\hat{X}_0^\alpha = \mathcal{D}\big(Z_t - \sigma_t\,\epsilon_{M_\theta^\alpha}(Z_t,X,P)\big)$ be the model's one-step clean-image estimate at scale $\alpha$, decoded to pixels, and let $D(\cdot)$ denote the embedding of a frozen DINOv2~\citep{caron2021emerging} feature encoder. We penalize
\vspace{-4pt}
\begin{equation}
\mathcal{L}_{\text{sem}} = \mathbb{E}\Big\| D\big(\hat{X}_0^{\alpha_m}\big) - \tfrac{1}{2}\big(\operatorname{sg}[D(\hat{X}_0^{\alpha_1})] + \operatorname{sg}[D(\hat{X}_0^{\alpha_2})]\big)\Big\|^2.
\label{eq:sem}
\vspace{-1pt}
\end{equation}
This directly regularizes the quantity a user cares about -- the perceived attribute -- rather than a latent proxy for it, and catches discontinuities that are invisible in velocity space but visible on screen.

\textbf{Spectral-norm regularizer.} SliderEdit reports needing early stopping to keep sliders smooth: continuing to train lets the adapter become extremely good, in raw loss terms, at suppressing the edit, but it does so by making the effective update $\Delta W^\ell = \alpha B^\ell A^\ell$ increasingly ``sharp''. Concretely, its spectral norm $\sigma_{\max}(\Delta W^\ell)$ grows, and since the adapter's contribution composes through every subsequent nonlinear layer, a large per-layer operator norm is precisely what turns a small change in $\alpha$ into a disproportionately large, discontinuous change in the output -- the network starts behaving like a hard switch rather than a gentle dial. Early stopping controls this only indirectly, by refusing to train long enough for it to happen, which caps edit strength along with it. We instead constrain it explicitly, penalizing the squared spectral norm of every layer's effective update: 
\vspace{-3pt}
\begin{equation}
\mathcal{L}_{\text{spec}} = \frac{1}{L}\sum_{\ell=1}^{L} \sigma_{\max}\!\big(\alpha\, B^\ell A^\ell\big)^2.
\label{eq:spec}
\vspace{-1pt}
\end{equation}
This allows training to full convergence while directly bounding adapter sensitivity to $\alpha$, replacing an ad-hoc heuristic with a principled constraint.

Overall, we train \modelname{} by jointly optimizing the base SliderEdit loss and our proposed regularizers: $\mathcal{L} = \mathcal{L}_{\text{PPS}} + \lambda_{\text{vel}}\mathcal{L}_{\text{vel}} + \lambda_{\text{sem}}\mathcal{L}_{\text{sem}} + \lambda_{\text{spec}}\mathcal{L}_{\text{spec}}$. The overall pipeline of our method is illustrated in Figure~\ref{fig:method} and Algorithm~\ref{alg:slidereditpp}.

\vspace{-4pt}
\subsection{Towards Solving Disentanglement Issues in Multi-Sliders}
\label{sec:disentangle}
\vspace{-4pt}

In multi-attribute editing, a composite prompt contains $K$ sub-instructions $P = \{P_1, \dots, P_K\}$ modulated by independent sliders $\boldsymbol{\alpha} = (\alpha_1, \dots, \alpha_K) \in [0,1]^K$. SliderEdit supports this via Selective-Token LoRA (STLoRA), which scales the adapter update at token positions corresponding to $P_i$ by $\alpha_i$ while optimizing a subset-suppression objective $\mathcal{L}_{\text{PPS}}^{\text{multi}}$ (see Appendix~\ref{sec:app:slideredit_multi_prelim} for full details). However, while token-level gating restricts \emph{where} the adapter is active, all instructions share the same LoRA weights and cross-attention mixing. Consequently, suppressing one instruction can inadvertently induce correlated output shifts along another, causing noticeable attribute leakage.

\textbf{Explicit disentanglement regularizer.} To prevent cross-attribute interference, we enforce that the velocity shifts induced by different sliders remain mutually orthogonal. For distinct instructions $a \neq b$, we define the individual effect direction of slider $i \in \{a, b\}$ relative to the unsuppressed baseline ($\boldsymbol{\alpha} = \mathbf{0}$) as:
\begin{equation}
\delta_i = \epsilon_{M_\theta(\mathbf{e}_i)}(Z_t, X, P) - \epsilon_{M_\theta(\mathbf{0})}(Z_t, X, P),
\label{eq:delta}
\end{equation}
where $\mathbf{e}_i$ denotes the one-hot scale vector isolating instruction $i$ at full suppression ($\alpha_i=1$). Geometrically, true disentanglement requires that isolated changes along attribute $a$ share no directional overlap with attribute $b$ ($\delta_a \perp \delta_b$). We enforce this condition directly by penalizing the absolute cosine similarity between effect directions:
\vspace{-3pt}
\begin{equation}
\mathcal{L}_{\text{dis}} = \mathbb{E}_{a \neq b}\left[ \frac{|\langle \delta_a, \delta_b \rangle|}{\|\delta_a\| \, \|\delta_b\|} \right].
\label{eq:dis}
\vspace{-2pt}
\end{equation}
Minimizing $\mathcal{L}_{\text{dis}}$ alongside $\mathcal{L}_{\text{PPS}}^{\text{multi}}$ pushes distinct editing trajectories toward orthogonality in velocity space, ensuring that modulating one slider alters the target attribute without leaking into others.

\vspace{-4pt}
\section{Experiments}
\vspace{-4pt}
We comprehensively evaluate \modelname{} on \benchmarkname{} through quantitative comparisons, qualitative results, and a human study. We compare against existing slider-based baselines, analyze the trade-off between controllability and inference speed, and ablate our proposed regularizers to isolate their contribution to smoothness and disentanglement.

\begin{figure}[t]
    \centering
    \includegraphics[width=0.94\textwidth]{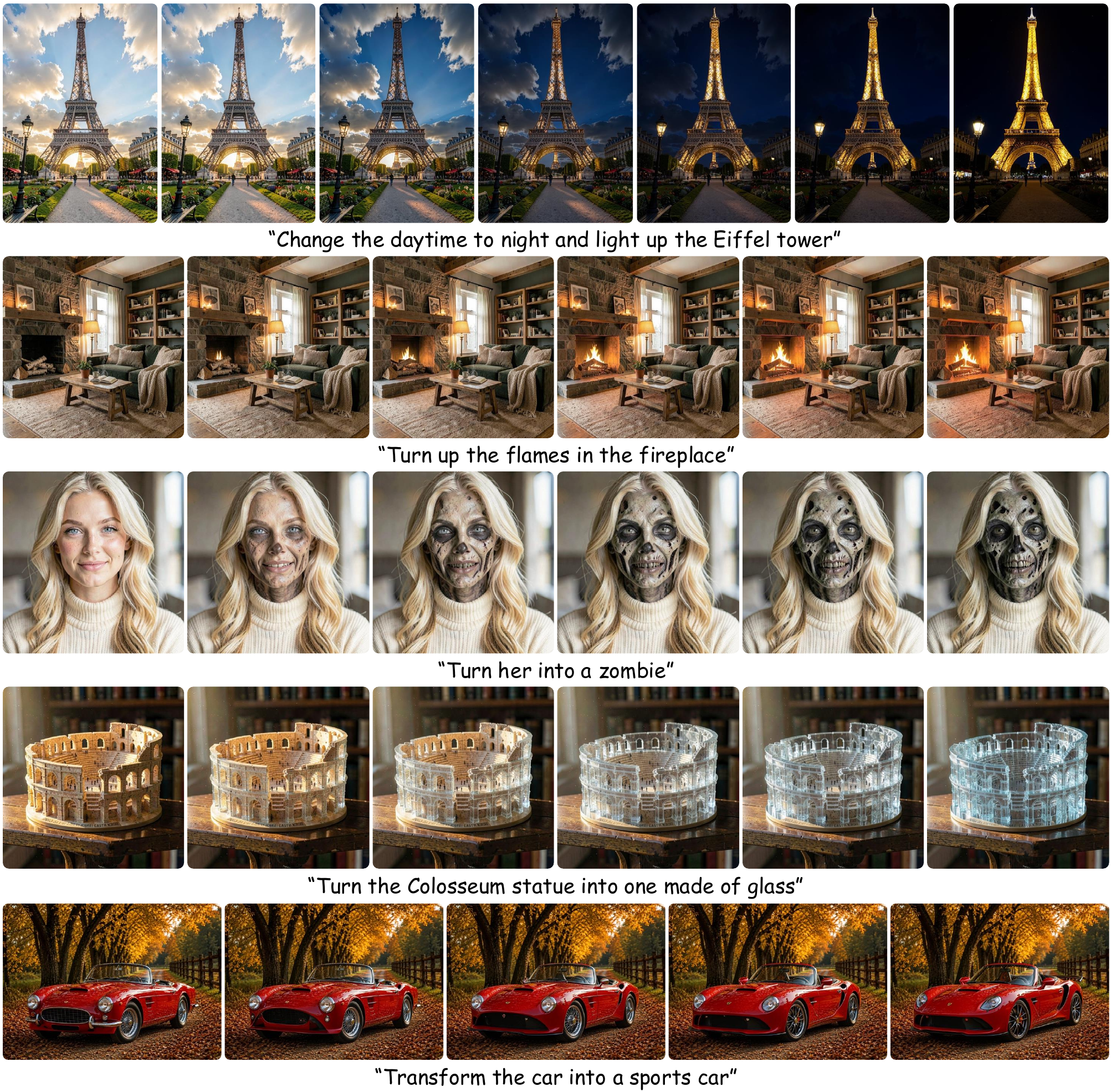}
    \vspace{-12pt}
    \caption{\textbf{\modelname{}} displays strong continuity across different transition levels of the slider. }
    \label{fig:qualitative_examples_general}
    \vspace{-14pt}
\end{figure}

\vspace{-4pt}
\subsection{Setting}
\vspace{-4pt}

We build \modelname{} on the FLUX-Klein 4B backbone, injecting low-rank adapters of rank 32 at every linear layer. All models are trained with the base $\mathcal{L}_{\text{PPS}}$ objective augmented with our regularizers, using loss weights $\lambda_{\mathrm{vel}} = 10.0$, $\lambda_{\mathrm{sem}} = 1.0$, and $\lambda_{\mathrm{spec}} = 0.1$; the semantic regularizer uses a frozen DINOv2-L encoder. We train for 1000 steps with Adam at learning rate $4\times10^{-5}$ and a global batch size of 32 on $8\times$H100. Unlike the baseline, we train to convergence rather than relying on early stopping, since the spectral-norm regularizer controls slider sharpness explicitly. At inference we use first-order Euler ODE solver with 8 denoising steps. For evaluation, each slider is rendered at $A = 8$ uniformly spaced positions $\alpha_i \in [0,1]$, and all metrics are averaged over the $285$ samples of the benchmark across its $7$ categories. We compare against KontinuousKontext~\citep{parihar2026kontinuous}, SDE-Edit~\citep{meng2021sdedit}, VeloEdit~\citep{li2026veloedit}, the original SliderEdit~\citep{zarei2026slideredit}, and our re-trained SliderEdit (Klein-4B) baseline, which applies the SliderEdit objective to the same backbone as ours and isolates the effect of the regularizers alone.

\subsection{Single Slider}
\begin{figure}[t]
    \centering
    \includegraphics[width=0.95\textwidth]{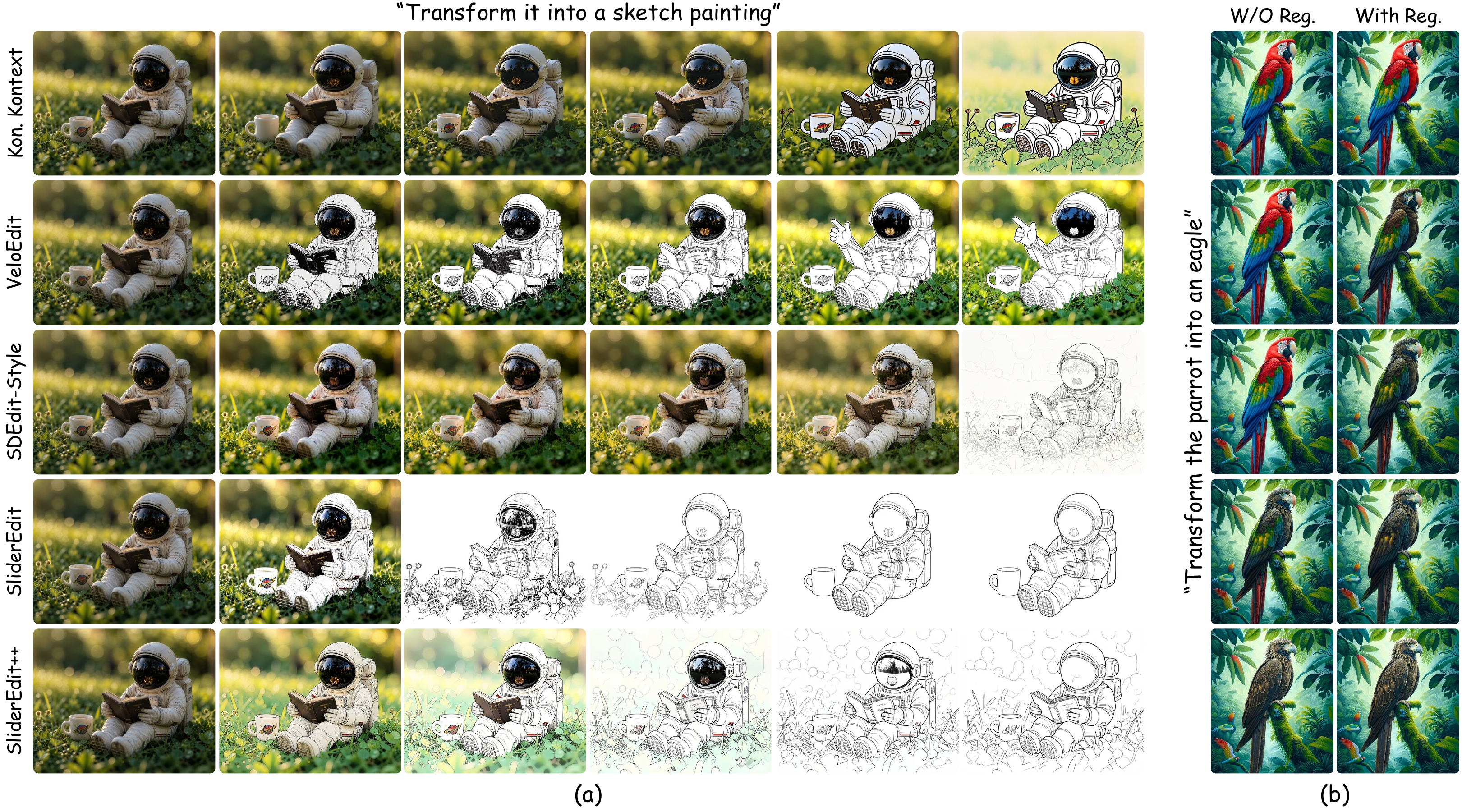}
    \vspace{-12pt}
    \caption{\textbf{Qualitative comparison with baselines and the effect of our proposed regularizers.} (a) demonstrates that \modelname{} produces smoother trajectories than competing baselines, while (b) illustrates the impact of our regularizers encouraging significantly smoother semantic transitions.}
    \label{fig:qualitative_comparison_baselines_and_without_regularizer}
\end{figure}

\begin{figure}[t]
\vspace{-5pt}
    \centering
    \begin{minipage}[t]{0.565\textwidth}
        \centering
        \vspace{0pt}
        \includegraphics[width=\linewidth]{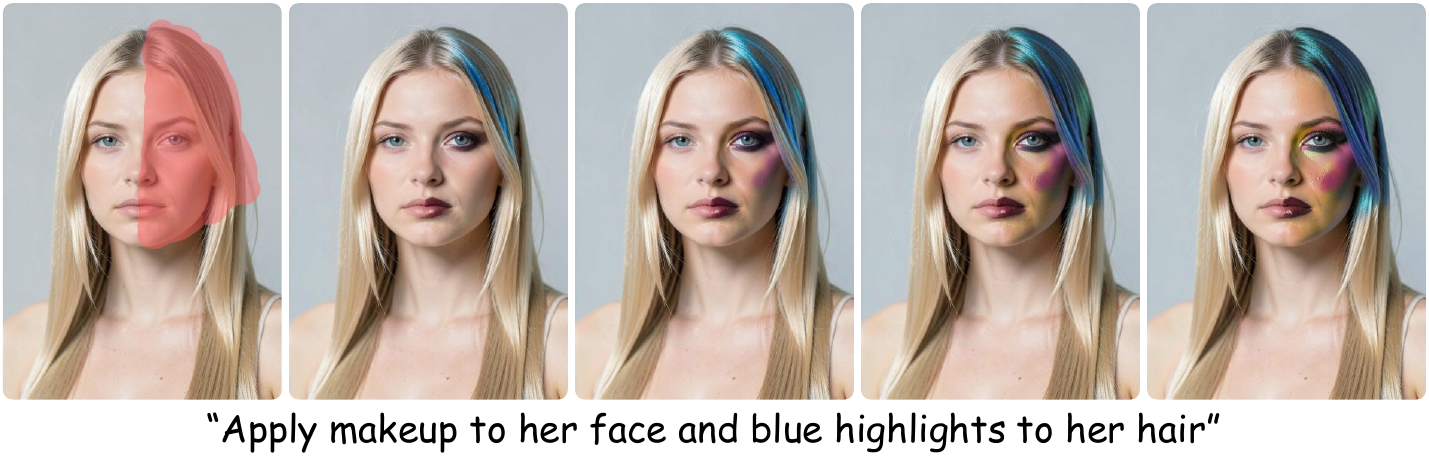}
        \vspace{-20pt}
        \caption{Extending \modelname{} to spatially localized editing, enabling continuous control within user-specified regions while preserving surrounding content.}
        \label{fig:masked_based_qualitative}
    \end{minipage}
    \hfill
    \begin{minipage}[t]{0.415\textwidth}
        \centering
        \vspace{0pt}
        \includegraphics[width=\linewidth]{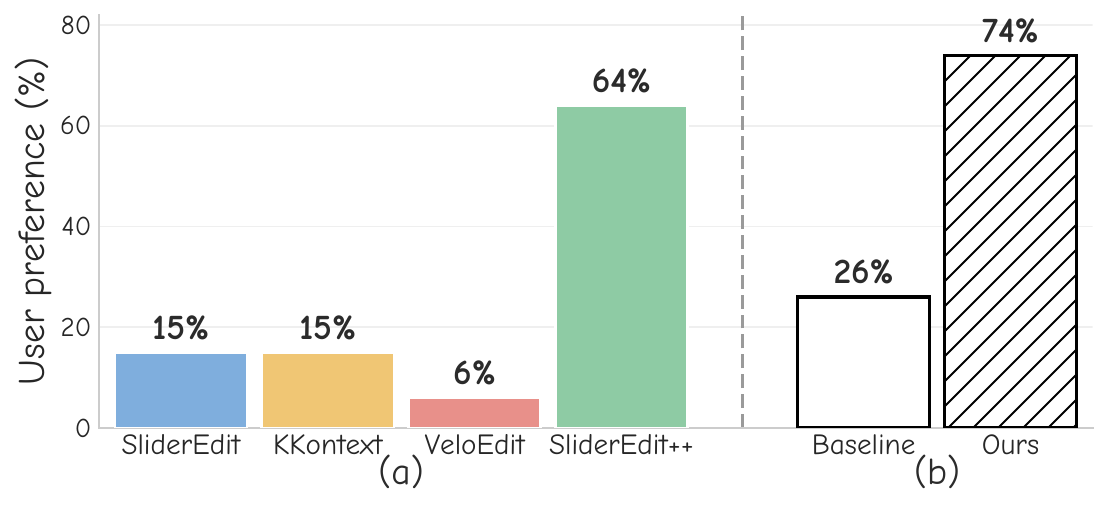}
        \vspace{-24pt}
        \caption{Human preference comparing our method with baselines (a) and our training strategies with the baseline (b).}
        \label{fig:human_prefrence_results}
    \end{minipage}
\end{figure}

\vspace{-6pt}
We first evaluate \modelname{} on \benchmarkname{} using the dataset and metrics from Section~\ref{sec:slidebench_evaluation}. Table~\ref{tab:baseline_metrics} compares against existing slider-based methods. It consistently outperforms all baselines in both monotonic progression and frame uniformity, yielding an overall 18.5\% relative improvement over the strongest prior method.  This advantage is also consistent across editing categories: Figure~\ref{fig:dataset_categories_and_models_performance} shows that \modelname{} achieves the best performance in every category of \benchmarkname{}. Importantly, this improved controllability does not come at the expense of efficiency; Figure~\ref{fig:performance_vs_speed_comparison} shows that \modelname{} achieves the best performance--latency trade-off, operating approximately 3$\times$ to 20$\times$ faster than competing slider methods. Qualitatively, Figure~\ref{fig:qualitative_comparison_baselines_and_without_regularizer}(a) shows that competing methods often exhibit abrupt visual changes along the traversal, whereas our method produces smoother and more gradual edit progression. Additional examples in Figure~\ref{fig:qualitative_examples_general} demonstrate this behavior across diverse global and localized edits, including lighting, material, object, and character transformations.
To validate that our automatic metrics reflect human perception, we conduct a human study where we ask annotators to select the smoothest and most edit-faithful slider sweep among randomized methods. Figure~\ref{fig:human_prefrence_results} (a) shows that \modelname{} outperforms all baselines by a substantial margin.

Furthermore, to isolate the effect of our proposed regularizers, we ablate them on the same backbone and training setup by enabling each term individually and in combination. As shown in Table~\ref{tab:regularizer_ablation}, every regularizer improves slider quality over the unregularized baseline, while combining all three yields the strongest performance, indicating that their effects are complementary. The qualitative comparisons in Figures~\ref{fig:qualitative_comparison_baselines_and_without_regularizer}(b) and~\ref{fig:qualitative_examples_with_and_without_regularizers_appendix_base} further illustrate this effect: without regularization, edits often emerge abruptly at isolated slider positions, whereas our full model introduces them progressively and smoothly across the trajectory. Finally, the human study in Figure~\ref{fig:human_prefrence_results} (b) shows a significant preference for our method over the baseline.

\begin{table}[t]
\vspace{-6pt}
\centering
\small
\setlength{\tabcolsep}{4pt}
\resizebox{\textwidth}{!}{%
\begin{tabular}{lccccccccc}
\toprule
 & \multicolumn{3}{c}{Monotonic Progression} & \multicolumn{5}{c}{Frame Uniformity × VLM Edit Strength} & \multicolumn{1}{c}{Overall} \\
\cmidrule(lr){2-4} \cmidrule(lr){5-9} \cmidrule(lr){10-10}
 & GPT-5 & Opus-4.8 & Avg & LPIPS & CLIP & SigLIP & DINOv2 & Avg & Score \\
\midrule
KontinuousKontext & \underline{0.7073} & 0.6992 & \underline{0.7033} & 0.6223 & 0.6626 & 0.6630 & 0.6585 & 0.6516 & 0.6774 \\
TextSteer & 0.4980 & 0.5120 & 0.5050 & 0.6020 & 0.6320 & 0.6510 & 0.6530 & 0.6345 & 0.5690 \\
SDE-Edit & 0.5363 & 0.5820 & 0.5591 & \underline{0.6528} & 0.6630 & 0.6634 & 0.6590 & 0.6596 & 0.6094 \\
VeloEdit & 0.6837 & \underline{0.7098} & 0.6967 & 0.6381 & 0.6845 & 0.6868 & 0.6816 & 0.6727 & \underline{0.6847} \\
Explicit CFG & 0.6381 & 0.6406 & 0.6393 & 0.6400 & 0.6795 & 0.6810 & 0.6746 & 0.6688 & 0.6541 \\
SliderEdit & 0.6135 & 0.6311 & 0.6223 & 0.6394 & \underline{0.6925} & \underline{0.6949} & \underline{0.6891} & \underline{0.6790} & 0.6506 \\
\midrule
Ours & \textbf{0.7801} & \textbf{0.7890} & \textbf{0.7845} & \textbf{0.8272} & \textbf{0.8431} & \textbf{0.8443} & \textbf{0.8399} & \textbf{0.8386} & \textbf{0.8116} \\
\rowcolor{gray!12}
\textit{\% gain} & \textcolor{ForestGreen}{+10.3\%} & \textcolor{ForestGreen}{+11.2\%} & \textcolor{ForestGreen}{+11.5\%} & \textcolor{ForestGreen}{+26.7\%} & \textcolor{ForestGreen}{+21.7\%} & \textcolor{ForestGreen}{+21.5\%} & \textcolor{ForestGreen}{+21.9\%} & \textcolor{ForestGreen}{+23.5\%} & \textcolor{ForestGreen}{+18.5\%} \\
\bottomrule
\end{tabular}
}
\vspace{-11pt}
\caption{Comparison on monotonic progression and frame uniformity. Best and second-best results are shown in \textbf{bold} and \underline{underline}; the final row reports our relative gain over the strongest baseline.}
\label{tab:baseline_metrics}
\end{table}
\begin{table}[!ht]
\vspace{-6pt}
\centering
\small
\begin{tabular}{lcccccc}
\toprule
Method & $\mathcal{L}_{\mathrm{sem}}$ & $\mathcal{L}_{\mathrm{vel}}$ & $\mathcal{L}_{\mathrm{spec}}$ & $MP_{avg}$ & $\bar U_{avg}$ & Overall \\
\midrule
Baseline & \xmark & \xmark & \xmark & 0.7161 & 0.7258 & 0.7209 \\
\noalign{\vskip 2pt}\hdashline\noalign{\vskip 2pt}
+ $\mathcal{L}_{\mathrm{sem}}$           & \cmark & \xmark & \xmark & 0.7643 & 0.8060 & 0.7851 \\
+ $\mathcal{L}_{\mathrm{vel}}$          & \xmark & \cmark & \xmark & 0.7552 & 0.7901 & 0.7726 \\
+ $\mathcal{L}_{\mathrm{spec}}$         & \xmark & \xmark & \cmark & 0.7421 & 0.7964 & 0.7692 \\
\noalign{\vskip 2pt}\hdashline\noalign{\vskip 2pt}
+ All Regularizers    & \cmark & \cmark & \cmark & \textbf{0.7845} & \textbf{0.8386} & \textbf{0.8116} \\
\bottomrule
\end{tabular}
\vspace{-8pt}
\caption{Effect of each regularizer on SliderEdit. $\mathcal{L}_{\mathrm{sem}}$: DINO-based semantic midpoint regularizer, $\mathcal{L}_{\mathrm{vel}}$  : latent velocity midpoint regularizer, $\mathcal{L}_{\mathrm{spec}}$: spectral norm regularizer on the LoRA matrices.}
\label{tab:regularizer_ablation}
\vspace{-5pt}
\end{table}

\begin{table}[t]
\vspace{-5pt}
\centering
\small
\setlength{\tabcolsep}{4pt}
\resizebox{\textwidth}{!}{%
\begin{tabular}{lccccccccccccc}
\toprule
 & \multicolumn{3}{c}{Monotonic Progression} & \multicolumn{5}{c}{Frame Uniformity × VLM Edit Strength} & \multicolumn{4}{c}{Disentanglement} & \multicolumn{1}{c}{Overall} \\
\cmidrule(lr){2-4} \cmidrule(lr){5-9} \cmidrule(lr){10-13} \cmidrule(lr){14-14}
 & GPT-5 & Opus-4.8 & Avg & LPIPS & CLIP & SigLIP & DINOv2 & Avg & CLIP & SigLIP & BLIP & Avg & Score \\
\midrule
SliderEdit & 0.6511 & 0.6687 & 0.6599 & 0.7294 & 0.7638 & 0.7708 & 0.7740 & 0.7595 & 0.9930 & 0.9939 & 0.9694 & 0.9854 & 0.7097 \\
\modelname{} & \textbf{0.7810} & \textbf{0.7918} & \textbf{0.7864} & \textbf{0.9683} & \textbf{0.9747} & \textbf{0.9782} & \textbf{0.9818} & \textbf{0.9757} & \textbf{0.9932} & \textbf{0.9957} & \textbf{0.9719} & \textbf{0.9869} & \textbf{0.8811} \\
\bottomrule
\end{tabular}
}
\vspace{-10pt}
\caption{Comparison of multi-slider baselines on monotonic progression, VLM-weighted frame uniformity, and disentanglement.}
\label{tab:multislider_quantitative_comparison}
\end{table}

\begin{figure}[!ht]
\vspace{-7pt}
    \centering
    \includegraphics[width=0.90\textwidth]{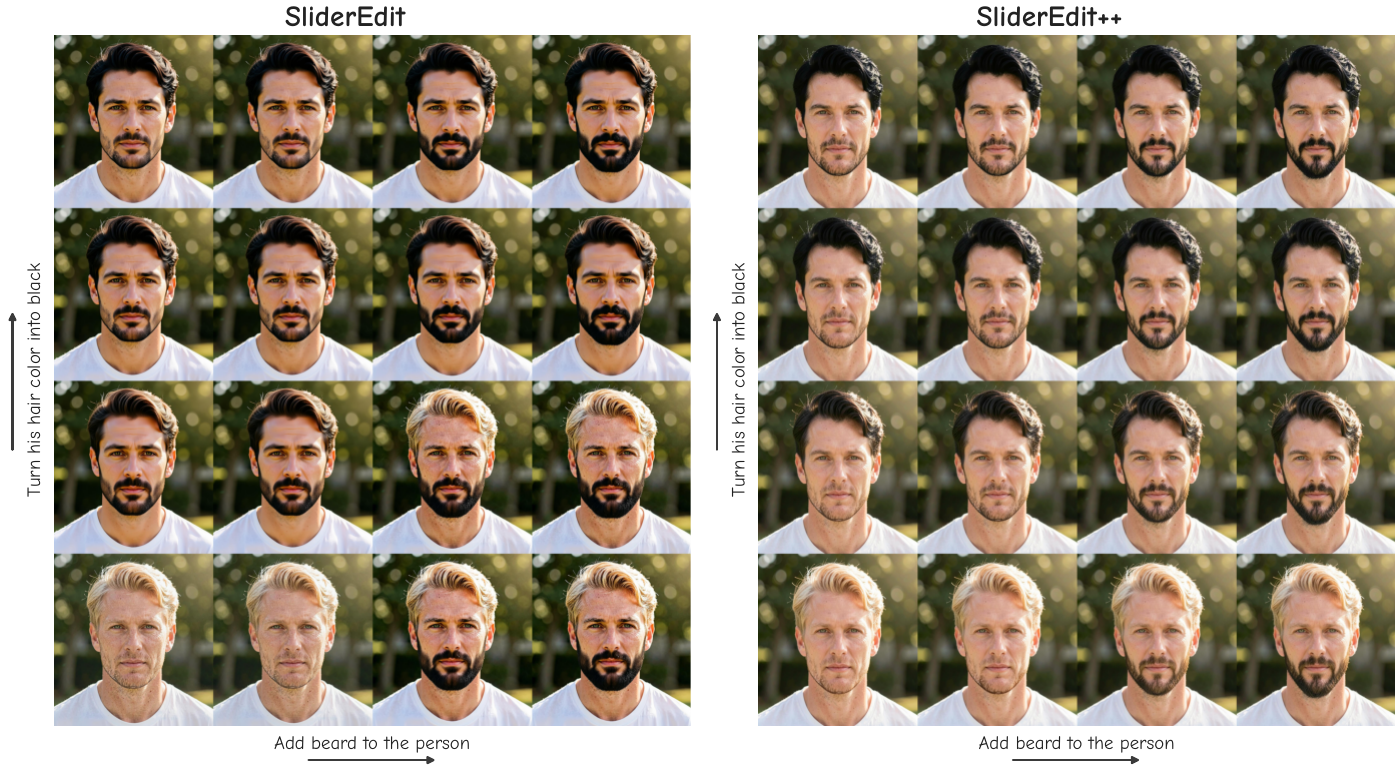}
    \vspace{-10pt}
    \caption{Qualitative results of \modelname{} showing strong disentanglement in multi-slider setting.}
    \label{fig:qualitative_examples_multi_slider}
\end{figure}

\vspace{-6pt}
\subsection{Multi Slider}
\vspace{-6pt}
We next evaluate the multi-slider setting, training \modelname{} with the multi-instruction $\mathcal{L}_{\mathrm{PPS}}$ objective together with our smoothness and disentanglement regularizers. Table~\ref{tab:multislider_quantitative_comparison} compares against the base SliderEdit model and shows consistent improvements across the evaluated metrics, increasing the overall score from $0.7097$ to $0.8811$. Figure~\ref{fig:qualitative_examples_multi_slider} further illustrates the difference: SliderEdit exhibits cross-slider interference, where adjusting one instruction can unintentionally affect another, whereas \modelname{} produces smoother and more disentangled controls, with changes to one slider having minimal effect on the others.

\vspace{-6pt}
\subsection{Extending to Mask-Conditioned Sliders}
\vspace{-6pt}
We extend \modelname{} to \emph{spatially localized} editing \citep{avrahami2022blended, zarei2026spatially}, enabling a
continuum of edits restricted to a user-specified region while the complement is
preserved. Given a source image $\mathbf{x}$, a binary mask
$\mathbf{m} \in \{0,1\}^{H\times W}$ (with $1$ denoting the region), and
an instruction, the edit magnitude is governed by the LoRA adapter scale $\alpha$.
To confine the edit, we adopt a masked latent-blending scheme: the model
performs a full-image edit, but during the final $k$ denoising steps we replace the
out-of-mask latents with a re-noised copy of the source,
\begin{equation}
  z_t \;\leftarrow\; \mathbf{m}\odot z_t \;+\;
  (1-\mathbf{m})\odot\!\big[(1-\sigma_t)\,z_0 + \sigma_t\,\boldsymbol{\epsilon}\big],
  \qquad \boldsymbol{\epsilon}\sim\mathcal{N}(\mathbf{0},\mathbf{I}),
  \label{eq:masked-blend}
\end{equation}
where $z_0$ is the (VAE-)encoded source and $\sigma_t$ the noise level at timestep
$t$, and $\odot$ denotes element-wise (spatially broadcast) multiplication.
Blending only the last $k$ steps (we use $k=2$) preserves a strong in-region edit
while yielding a smooth latent boundary. Figure~\ref{fig:masked_based_qualitative} shows a qualitative example in which the user specifies an edit region and instruction, and \modelname{} provides continuous control over the localized edit.
We provide extensive qualitative results in Figure~\ref{fig:masked_conditioned_qualitative_appendix}.

\section{Conclusion}

In this paper, we first introduced \benchmarkname{}, a comprehensive benchmark for evaluating continuous image editing, and showed that existing slider-based methods often suffer from discontinuous trajectories, abrupt visual jumps, and cross-slider interference. We then proposed \modelname{}, which introduces complementary regularizers to explicitly improve slider smoothness, uniform progression, and disentanglement during training. On \benchmarkname{}, \modelname{} achieves state-of-the-art performance, improving the overall score by 18.5\% over the strongest baseline while operating approximately 3$\times$ to 20$\times$ faster, enabling more reliable and responsive continuous image editing.

\bibliography{iclr2027_conference}

@inproceedings{zarei2026slideredit,
  title={Slideredit: Continuous image editing with fine-grained instruction control},
  author={Zarei, Arman and Basu, Samyadeep and Pournemat, Mobina and Nag, Sayan and Rossi, Ryan A and Feizi, Soheil},
  booktitle={Proceedings of the IEEE/CVF Conference on Computer Vision and Pattern Recognition},
  pages={14430--14439},
  year={2026}
}

@inproceedings{parihar2026kontinuous,
  title={Kontinuous kontext: Continuous strength control for instruction-based image editing},
  author={Parihar, Rishubh and Patashnik, Or and Ostashev, Daniil and Radhakrishnan, Venkatesh Babu and Cohen-Or, Daniel and Wang, Kuan-Chieh Jackson},
  booktitle={Proceedings of the IEEE/CVF Conference on Computer Vision and Pattern Recognition},
  pages={37929--37939},
  year={2026}
}

@article{li2026veloedit,
  title={VeloEdit: Training-Free Consistent and Continuous Instruction-Based Image Editing via Velocity Field Decomposition},
  author={Li, Zongqing and Liu, Zhihui and Xie, Yujie and Wu, Shansiyuan and Lv, Hongshen and Su, Songzhi},
  year={2026},
  publisher={TechRxiv}
}

@article{ekin2026unreasonable,
  title={The unreasonable effectiveness of text embedding interpolation for continuous image steering},
  author={Ekin, Yigit and Gandelsman, Yossi},
  journal={arXiv preprint arXiv:2603.17998},
  year={2026}
}

@article{meng2021sdedit,
  title={Sdedit: Guided image synthesis and editing with stochastic differential equations},
  author={Meng, Chenlin and He, Yutong and Song, Yang and Song, Jiaming and Wu, Jiajun and Zhu, Jun-Yan and Ermon, Stefano},
  journal={arXiv preprint arXiv:2108.01073},
  year={2021}
}

@inproceedings{caron2021emerging,
  title={Emerging properties in self-supervised vision transformers},
  author={Caron, Mathilde and Touvron, Hugo and Misra, Ishan and J{\'e}gou, Herv{\'e} and Mairal, Julien and Bojanowski, Piotr and Joulin, Armand},
  booktitle={2021 IEEE/CVF international conference on computer vision (ICCV)},
  pages={9630--9640},
  year={2021},
  organization={IEEE}
}

@article{ho2022classifier,
  title={Classifier-free diffusion guidance},
  author={Ho, Jonathan and Salimans, Tim},
  journal={arXiv preprint arXiv:2207.12598},
  year={2022}
}

@article{labs2025flux,
  title={Flux. 1 kontext: Flow matching for in-context image generation and editing in latent space},
  author={Labs, Black Forest and Batifol, Stephen and Blattmann, Andreas and Boesel, Frederic and Consul, Saksham and Diagne, Cyril and Dockhorn, Tim and English, Jack and English, Zion and Esser, Patrick and others},
  journal={arXiv preprint arXiv:2506.15742},
  year={2025}
}

@article{wu2025qwen,
  title={Qwen-image technical report},
  author={Wu, Chenfei and Li, Jiahao and Zhou, Jingren and Lin, Junyang and Gao, Kaiyuan and Yan, Kun and Yin, Sheng-ming and Bai, Shuai and Xu, Xiao and Chen, Yilei and others},
  journal={arXiv preprint arXiv:2508.02324},
  year={2025}
}

@misc{flux-2-2025,
    author={Black Forest Labs},
    title={{FLUX.2: Frontier Visual Intelligence}},
    year={2025},
    howpublished={\url{https://bfl.ai/blog/flux-2}},
}

@inproceedings{
higgins2017betavae,
title={beta-{VAE}: Learning Basic Visual Concepts with a Constrained Variational Framework},
author={Irina Higgins and Loic Matthey and Arka Pal and Christopher Burgess and Xavier Glorot and Matthew Botvinick and Shakir Mohamed and Alexander Lerchner},
booktitle={International Conference on Learning Representations},
year={2017},
url={https://openreview.net/forum?id=Sy2fzU9gl}
}

@inproceedings{kim2018disentangling,
  title={Disentangling by factorising},
  author={Kim, Hyunjik and Mnih, Andriy},
  booktitle={International conference on machine learning},
  pages={2649--2658},
  year={2018},
  organization={PMLR}
}

@article{shen2020interfacegan,
  title={Interfacegan: Interpreting the disentangled face representation learned by gans},
  author={Shen, Yujun and Yang, Ceyuan and Tang, Xiaoou and Zhou, Bolei},
  journal={IEEE transactions on pattern analysis and machine intelligence},
  volume={44},
  number={4},
  pages={2004--2018},
  year={2020},
  publisher={IEEE}
}

@article{harkonen2020ganspace,
  title={Ganspace: Discovering interpretable gan controls},
  author={H{\"a}rk{\"o}nen, Erik and Hertzmann, Aaron and Lehtinen, Jaakko and Paris, Sylvain},
  journal={Advances in neural information processing systems},
  volume={33},
  pages={9841--9850},
  year={2020}
}

@inproceedings{shen2021closed,
  title={Closed-form factorization of latent semantics in gans},
  author={Shen, Yujun and Zhou, Bolei},
  booktitle={2021 IEEE/CVF Conference on Computer Vision and Pattern Recognition (CVPR)},
  pages={1532--1540},
  year={2021},
  organization={IEEE}
}

@article{abdal2021styleflow,
  title={Styleflow: Attribute-conditioned exploration of stylegan-generated images using conditional continuous normalizing flows},
  author={Abdal, Rameen and Zhu, Peihao and Mitra, Niloy J and Wonka, Peter},
  journal={ACM Transactions on Graphics (ToG)},
  volume={40},
  number={3},
  pages={1--21},
  year={2021},
  publisher={ACM New York, NY}
}

@inproceedings{gandikota2024concept,
  title={Concept sliders: Lora adaptors for precise control in diffusion models},
  author={Gandikota, Rohit and Materzy{\'n}ska, Joanna and Zhou, Tingrui and Torralba, Antonio and Bau, David},
  booktitle={European Conference on Computer Vision},
  pages={172--188},
  year={2024},
  organization={Springer}
}

@inproceedings{sridhar2024prompt,
  title={Prompt sliders for fine-grained control, editing and erasing of concepts in diffusion models},
  author={Sridhar, Deepak and Vasconcelos, Nuno},
  booktitle={European Conference on Computer Vision},
  pages={17--31},
  year={2024},
  organization={Springer}
}

@inproceedings{baumann2025continuous,
  title={Continuous, subject-specific attribute control in t2i models by identifying semantic directions},
  author={Baumann, Stefan Andreas and Krause, Felix and Neumayr, Michael and Stracke, Nick and Sevi, Melvin and Hu, Vincent Tao and Ommer, Bj{\"o}rn},
  booktitle={2025 IEEE/CVF Conference on Computer Vision and Pattern Recognition (CVPR)},
  pages={13231--13241},
  year={2025},
  organization={IEEE}
}

@inproceedings{gandikota2025sliderspace,
  title={Sliderspace: Decomposing the visual capabilities of diffusion models},
  author={Gandikota, Rohit and Wu, Zongze and Zhang, Richard and Bau, David and Shechtman, Eli and Kolkin, Nick},
  booktitle={2025 IEEE/CVF International Conference on Computer Vision (ICCV)},
  pages={15994--16003},
  year={2025},
  organization={IEEE}
}

@inproceedings{chiu2026text,
  title={Text Slider: Efficient and Plug-and-Play Continuous Concept Control for Image/Video Synthesis via LoRA Adapters},
  author={Chiu, Pin-Yen and Fang, I and Chen, Jun-Cheng and others},
  booktitle={Proceedings of the IEEE/CVF Winter Conference on Applications of Computer Vision},
  pages={613--622},
  year={2026}
}

@article{dalva2024fluxspace,
  title={Fluxspace: Disentangled semantic editing in rectified flow transformers},
  author={Dalva, Yusuf and Venkatesh, Kavana and Yanardag, Pinar},
  journal={arXiv preprint arXiv:2412.09611},
  year={2024}
}

@article{endo2026flowslider,
  title={FlowSlider: Training-Free Continuous Image Editing via Fidelity-Steering Decomposition},
  author={Endo, Taichi and Hao, Guoqing and Sumi, Kazuhiko},
  journal={arXiv preprint arXiv:2604.02088},
  year={2026}
}

@inproceedings{radford2021learning,
  title={Learning transferable visual models from natural language supervision},
  author={Radford, Alec and Kim, Jong Wook and Hallacy, Chris and Ramesh, Aditya and Goh, Gabriel and Agarwal, Sandhini and Sastry, Girish and Askell, Amanda and Mishkin, Pamela and Clark, Jack and others},
  booktitle={International conference on machine learning},
  pages={8748--8763},
  year={2021},
  organization={PmLR}
}

@inproceedings{li2022blip,
  title={Blip: Bootstrapping language-image pre-training for unified vision-language understanding and generation},
  author={Li, Junnan and Li, Dongxu and Xiong, Caiming and Hoi, Steven},
  booktitle={International conference on machine learning},
  pages={12888--12900},
  year={2022},
  organization={PMLR}
}

@inproceedings{zhai2023sigmoid,
  title={Sigmoid loss for language image pre-training},
  author={Zhai, Xiaohua and Mustafa, Basil and Kolesnikov, Alexander and Beyer, Lucas},
  booktitle={2023 IEEE/CVF International Conference on Computer Vision (ICCV)},
  pages={11941--11952},
  year={2023},
  organization={IEEE}
}

@article{singh2025openai,
  title={Openai gpt-5 system card},
  author={Singh, Aaditya and Fry, Adam and Perelman, Adam and Tart, Adam and Ganesh, Adi and El-Kishky, Ahmed and McLaughlin, Aidan and Low, Aiden and Ostrow, AJ and Ananthram, Akhila and others},
  journal={arXiv preprint arXiv:2601.03267},
  year={2025}
}

@article{liu2023visual,
  title={Visual instruction tuning},
  author={Liu, Haotian and Li, Chunyuan and Wu, Qingyang and Lee, Yong Jae},
  journal={Advances in neural information processing systems},
  volume={36},
  pages={34892--34916},
  year={2023}
}

@article{bai2023qwen,
  title={Qwen technical report},
  author={Bai, Jinze and Bai, Shuai and Chu, Yunfei and Cui, Zeyu and Dang, Kai and Deng, Xiaodong and Fan, Yang and Ge, Wenbin and Han, Yu and Huang, Fei and others},
  journal={arXiv preprint arXiv:2309.16609},
  year={2023}
}

@inproceedings{zhang2018unreasonable,
  title={The unreasonable effectiveness of deep features as a perceptual metric},
  author={Zhang, Richard and Isola, Phillip and Efros, Alexei A and Shechtman, Eli and Wang, Oliver},
  booktitle={2018 IEEE/CVF conference on computer vision and pattern recognition},
  pages={586--595},
  year={2018},
  organization={IEEE}
}

@article{zarei2026spatially,
  title={Spatially-Grounded Flow Matching: Structured Source Distributions for Image Generation},
  author={Zarei, Arman and Kalayeh, Mahdi M},
  journal={arXiv preprint arXiv:2608.15452},
  year={2026}
}

@inproceedings{avrahami2022blended,
  title={Blended diffusion for text-driven editing of natural images},
  author={Avrahami, Omri and Lischinski, Dani and Fried, Ohad},
  booktitle={2022 IEEE/CVF Conference on Computer Vision and Pattern Recognition (CVPR)},
  pages={18187--18197},
  year={2022},
  organization={IEEE}
}
\bibliographystyle{iclr2027_conference}

\newpage
\appendix
\definecolor{ppnew}{HTML}{000000}
\definecolor{ppbase}{gray}{0.52}
\begin{figure}[t]
\centering
\begin{minipage}{0.8\linewidth}
\begin{algorithm}[H]
\caption{Training SliderEdit++ (SliderEdit in \textcolor{ppbase}{gray}, our additions in \textcolor{ppnew}{black})}
\label{alg:slidereditpp}
\begin{algorithmic}[1]
\color{ppbase}
\Require $\epsilon(Z, X, P)$: Image Editing Model, $M_\theta$: Trainable Adapter, $\{(X^{(i)}_\text{orig}, \mathcal{P}^{(i)})\}$: Dataset, \textcolor{ppnew}{$\phi$: DINOv2 encoder, $\lambda_\text{vel},\lambda_\text{sem},\lambda_\text{spec},\lambda_\text{dis}$}
\For{each training step}
    \State $X_\text{orig}, \mathcal{P}=\{\mathcal{P}_1,\ldots,\mathcal{P}_K\}$ $\leftarrow$ Sample a Data
    \State $\varepsilon \sim \mathcal{N}(0,I), t \sim \mathcal{U}[0,1]$
    \If{use $\mathcal{L}_\text{SPPS}$}
        \State $\mathcal{P} = \{\mathcal{P}'_1\}$
    \EndIf
    \State $Z \leftarrow (1-t)\varepsilon + tX_\text{orig}$
    \State $\mathcal{P}_i \leftarrow$ Random target instruction from $\mathcal{P}$ to suppress
    \State $v^\star \leftarrow \epsilon(Z, X_\text{orig}, \mathcal{P}\setminus\{\mathcal{P}_i\})$
    \State $\hat{v} \leftarrow \epsilon_{M^{1}_\theta(\mathcal{P}_i)}(Z, X_\text{orig}, \mathcal{P})$
    \State $\mathcal{L}_\texttt{PPS} = \|\hat{v} - v^\star\|^2$
    \color{ppnew}
    \State $\alpha_1,\alpha_2 \sim \mathcal{U}[0,1]$, \; $\alpha_m \leftarrow \tfrac{1}{2}(\alpha_1+\alpha_2)$
    \State $v^{\alpha} \leftarrow \epsilon_{M^{\alpha}_\theta(\mathcal{P}_i)}(Z, X_\text{orig}, \mathcal{P})$ for $\alpha\in\{\alpha_1,\alpha_2,\alpha_m\}$
    \State $f^{\alpha} \leftarrow \phi\big(\hat{X}_0(Z, v^{\alpha})\big)$
    \State $\mathcal{L}_\text{vel} = \big\|v^{\alpha_m} - \tfrac{1}{2}(\mathrm{sg}[v^{\alpha_1}]+\mathrm{sg}[v^{\alpha_2}])\big\|^2$
    \State $\mathcal{L}_\text{sem} = \big\|f^{\alpha_m} - \tfrac{1}{2}(\mathrm{sg}[f^{\alpha_1}]+\mathrm{sg}[f^{\alpha_2}])\big\|^2$
    \State $\mathcal{L}_\text{spec} = \frac{1}{L}\sum_{\ell} \sigma_\text{max}(B^\ell A^\ell)^2$
    \State $\mathcal{L} = \mathcal{L}_\texttt{PPS} + \lambda_\text{vel}\mathcal{L}_\text{vel} + \lambda_\text{sem}\mathcal{L}_\text{sem} + \lambda_\text{spec}\mathcal{L}_\text{spec}$
    \If{$K \geq 2$}
        \State $\mathcal{P}_j \leftarrow$ Random instruction from $\mathcal{P}\setminus\{\mathcal{P}_i\}$
        \State $\delta_k \leftarrow \epsilon_{M^{1}_\theta(\mathcal{P}_k)}(Z, X_\text{orig}, \mathcal{P}) - \epsilon(Z, X_\text{orig}, \mathcal{P})$ for $k\in\{i,j\}$
        \State $\mathcal{L}_\text{dis} = |\langle\delta_i,\delta_j\rangle| \,/\, (\|\delta_i\|\|\delta_j\|)$
        \State $\mathcal{L} \mathrel{+}= \lambda_\text{dis}\mathcal{L}_\text{dis}$
    \EndIf
    \color{ppbase}
    \State Update $\theta$ via gradient descent on $\mathcal{L}$
\EndFor
\vspace{2pt}
\end{algorithmic}
\end{algorithm}
\end{minipage}
\end{figure}

\section{Appendix}

\section{Preliminaries}

\subsection{Continuous Multi-Slider Edit using SliderEdit}
\label{sec:app:slideredit_multi_prelim}

In the multi-attribute setting, a single edit instruction decomposes into $K$ independent sub-instructions, $P = \{P_1,\dots,P_K\}$, each with its own slider value $\alpha_i\in[0,1]$ (e.g.\ ``add a smile'' and ``make the hair curlier'' issued together). The natural extension of Section~\ref{sec:continuity}'s single global $\alpha$ is to let the adapter's contribution at each layer depend on \emph{token position}: at a token belonging to instruction $P_i$'s span, the effective update is $\alpha_i\,\Delta W^\ell$; elsewhere it is zero. This is the Selective-Token LoRA (STLoRA) mechanism: one shared set of weights realizes $K$ independently addressable sliders in a single forward pass, and the training loss is the same suppression objective as Eq.~\eqref{eq:pps}, now applied per (randomly chosen) subset of suppressed instructions $\mathcal{S}\subseteq\{1,\dots,K\}$, with the target computed from the base model on the prompt with exactly those instructions actually removed, $P\setminus\mathcal{S}$:
\begin{equation}
\mathcal{L}_{\text{PPS}}^{\text{multi}} = \mathbb{E}\Big\| \epsilon_{M_\theta(\mathcal{S})}(Z_t,X,P) - \epsilon(Z_t,X,P\setminus\mathcal{S}) \Big\|^2 .
\label{eq:pps-multi}
\end{equation}
Confining each instruction's LoRA activation to its own tokens is already a strong structural bias toward disentanglement -- attribute $i$ is only ever edited through its own tokens -- and inherits the same continuity treatment as the single-slider case at the per-instruction level.

\section{SlideBench}
\subsection{Dataset}
\vspace{-4pt}
\label{sec:app:eval_dataset}
To systematically evaluate continuous image editing, we construct a benchmark spanning a broad range of attributes that users may wish to adjust continuously in practice. The dataset is organized into seven high-level categories---\emph{Lighting}, \emph{Composition}, \emph{Camera}, \emph{Environment}, \emph{Character}, \emph{Style}, and \emph{Object}---covering $57$ fine-grained subcategories, with five samples per subcategory. Each subcategory defines a distinct continuous editing axis, such as \emph{time of day}, \emph{crowd density}, \emph{smile intensity}, \emph{color saturation}, or \emph{object aging}. Formally, each sample is represented as a triplet $(c, I, p)$, where $c$ is a source caption, $I$ is the corresponding source image, and $p$ is the editing instruction. The caption specifies one endpoint of the editing axis and is used to generate the source image, while the instruction defines the desired transformation toward the opposite endpoint. For example, a source caption may describe ``a village street photographed at golden hour with long warm shadows'', while the corresponding instruction requests ``changing the scene to midnight with a dark sky and dim street lamps''. Within each subcategory, we instantiate the same editing axis across diverse visual contexts to evaluate whether continuous control generalizes beyond a particular scene or image.

The benchmark is designed to capture qualitatively different forms of continuous control. Attributes in \emph{Lighting}, \emph{Style}, and \emph{Camera} naturally admit gradual variation and therefore directly test smooth and monotonic editing behavior. In contrast, \emph{Environment} and \emph{Object} edits, such as crowd density, clutter, aging, and corrosion, require models to progressively introduce or strengthen visual content rather than transition through abrupt discrete changes. \emph{Character}, our largest category with $21$ subcategories, further evaluates fine-grained control over facial and bodily attributes, where even small variations can be perceptually significant. Together, these categories provide broad coverage of continuous editing behavior and expose where existing methods succeed or fail across different types of control. 

\vspace{-4pt}
\subsection{Metrics}
\vspace{-4pt}
For evaluating \emph{monotonic progression}, we initially explored text-image similarity models such as CLIP~\cite{radford2021learning}, BLIP~\cite{li2022blip}, and SigLIP~\cite{zhai2023sigmoid}. However, we found these models insufficiently sensitive to the fine-grained differences that arise along an editing trajectory. In particular, small changes in attribute strength often produce noisy or inconsistent similarity scores, making these models unreliable for assessing whether an edit progresses smoothly and monotonically. We therefore rely on strong vision-language models (VLMs), following the evaluation protocol described in Section~\ref{sec:evaluation_metrics}. We further observed that highly capable VLMs, such as GPT-5~\cite{singh2025openai} and Claude Opus 4.8, provide substantially more consistent assessments than smaller open-source alternatives such as LLaVA~\cite{liu2023visual} and Qwen~\cite{bai2023qwen}. We consequently use these stronger models for our main evaluation.

For \emph{frame uniformity}, we measure perceptual and semantic changes between consecutive frames using LPIPS~\cite{zhang2018unreasonable}, CLIP~\cite{radford2021learning}, SigLIP~\cite{zhai2023sigmoid}, and DINOv2~\cite{caron2021emerging}. For \emph{edit strength}, we again use the strong VLM-based evaluator described above, and report the aggregated metric to obtain a more robust assessment.

Finally, \emph{disentanglement} is measured in the feature spaces of CLIP, SigLIP, and BLIP. While these metrics provide a useful proxy for unintended semantic changes, they remain limited by the same lack of sensitivity to fine-grained edits discussed above and should therefore not be viewed as a fully reliable measure of disentanglement. We complement them with human evaluation, while developing more robust feature spaces for capturing such granular editing effects remains an important direction for future work.

\begin{table}[t]
\centering
\small
\setlength{\tabcolsep}{4pt}
\resizebox{\textwidth}{!}{%
\begin{tabular}{lccccccccc}
\toprule
 & \multicolumn{3}{c}{Monotonic Progression} & \multicolumn{5}{c}{Frame Uniformity × VLM Edit Strength} & \multicolumn{1}{c}{Overall} \\
\cmidrule(lr){2-4} \cmidrule(lr){5-9} \cmidrule(lr){10-10}
 & GPT-5 & Opus-4.8 & Avg & LPIPS & CLIP & SigLIP & DINOv2 & Avg & Score \\
\midrule
$T=8$ & \underline{0.7589} & \textbf{0.8000} & \textbf{0.7794} & 0.7977 & 0.8130 & 0.8146 & 0.8104 & 0.8089 & 0.7942 \\
$T=4$ & \textbf{0.7629} & \underline{0.7950} & \underline{0.7789} & 0.8141 & 0.8275 & 0.8287 & 0.8244 & 0.8237 & \textbf{0.8013} \\
$T=2$ & 0.7398 & 0.7845 & 0.7622 & \underline{0.8291} & \textbf{0.8396} & \textbf{0.8411} & \textbf{0.8378} & \textbf{0.8369} & \underline{0.7995} \\
$T=1$ & 0.6937 & 0.7639 & 0.7288 & \textbf{0.8304} & \underline{0.8369} & \underline{0.8387} & \underline{0.8361} & \underline{0.8355} & 0.7822 \\
\bottomrule
\end{tabular}
}
\vspace{-12pt}
\caption{\textbf{Ablating the number of generation steps for slider quality.} Results show that using fewer generation steps has minimal impact on slider performance while significantly improving generation speed.}
\label{tab:ablating_denoising_steps}
\end{table}

\vspace{-4pt}
\section{Experiments}
\label{sec:app:experiments}
\vspace{-4pt}

\subsection{Results}
\begin{figure}[t]
    \centering
    \includegraphics[width=0.80\textwidth]{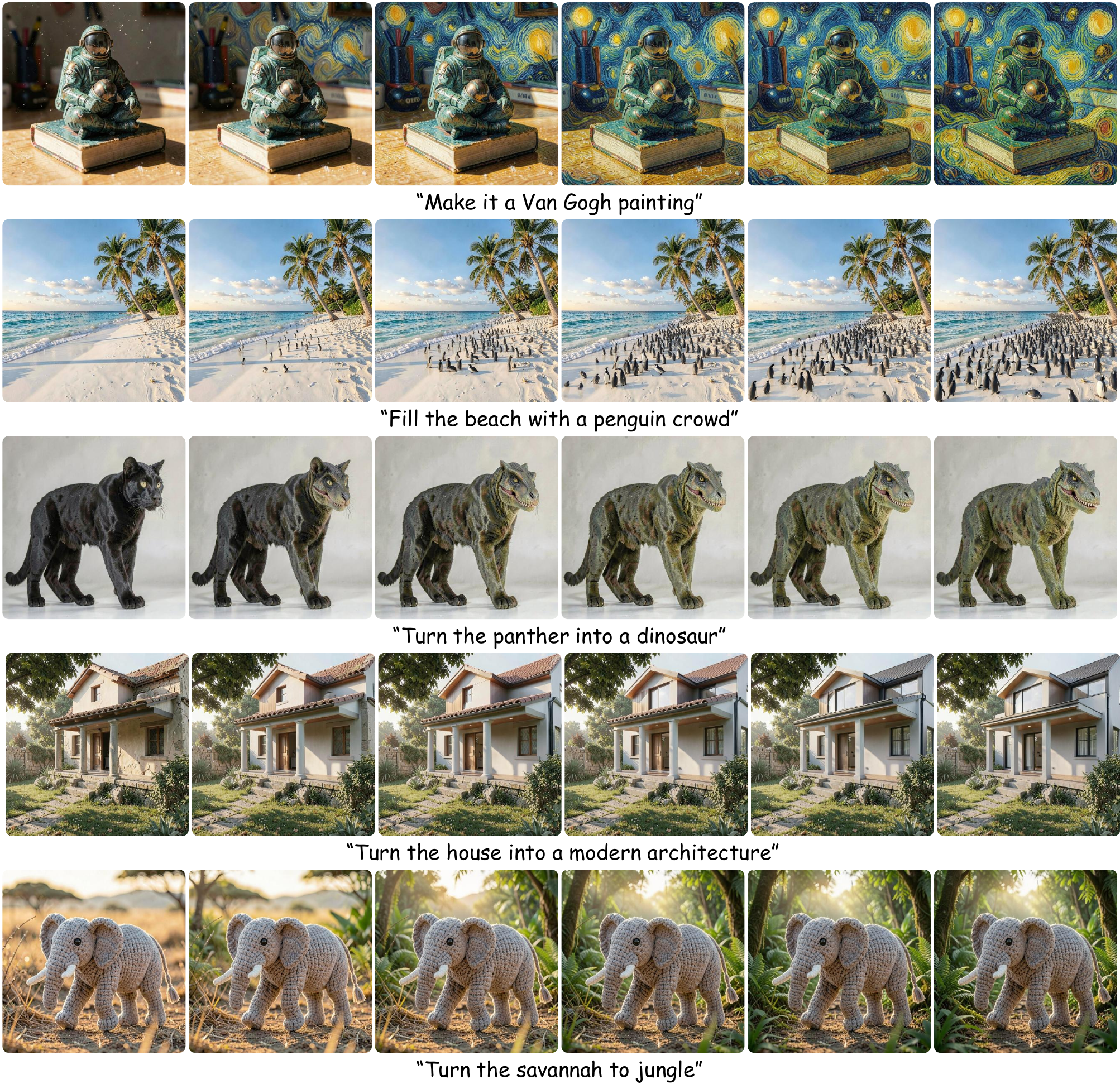}
    \includegraphics[width=0.80\textwidth]{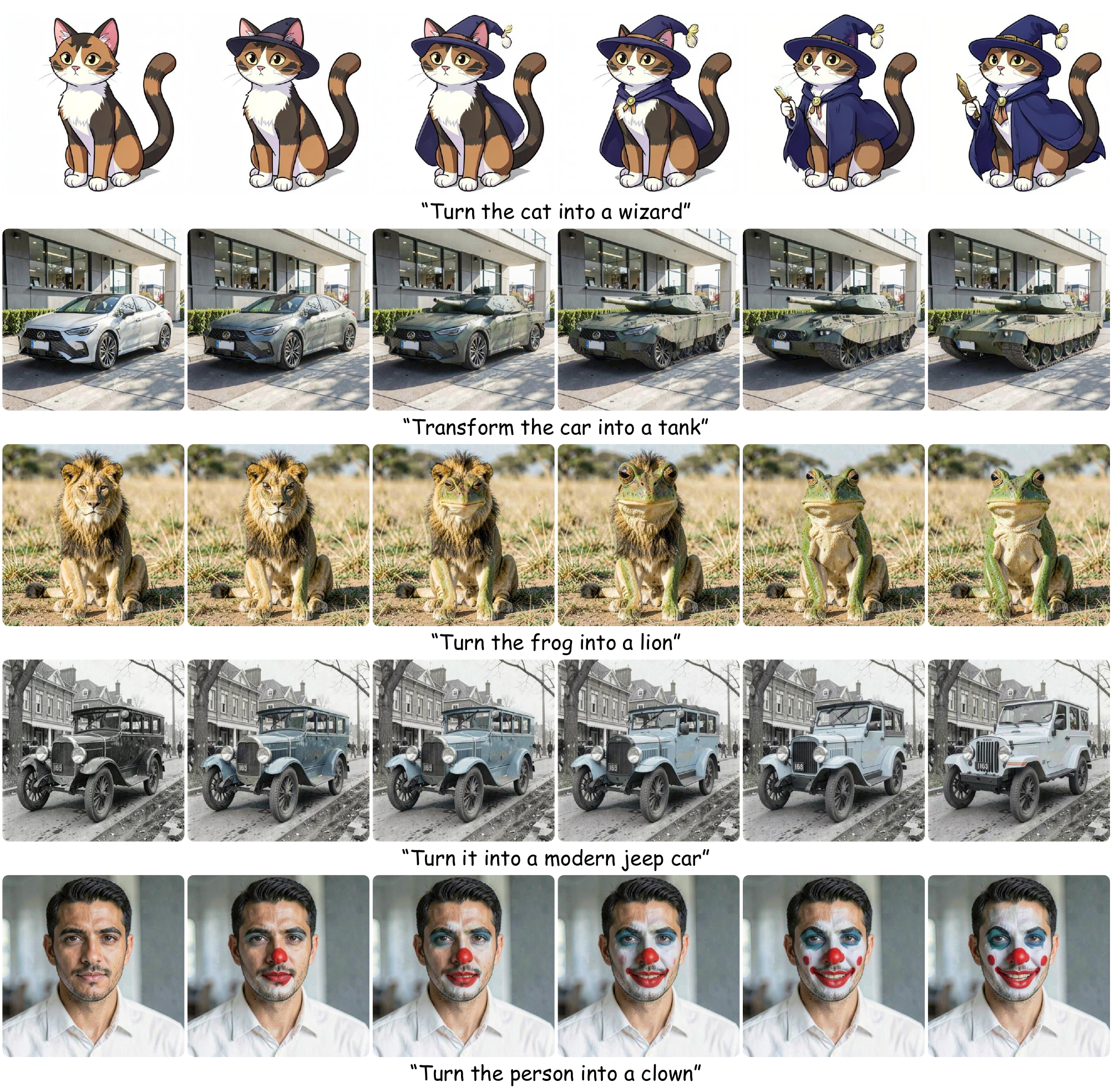}
    \vspace{-12pt}
    \caption{\modelname{} displays strong continuity across different transition levels of the slider. }
    \label{fig:qualitative_examples_appendix}
\end{figure}

\begin{figure}[t]
    \centering
    \includegraphics[width=0.95\textwidth]{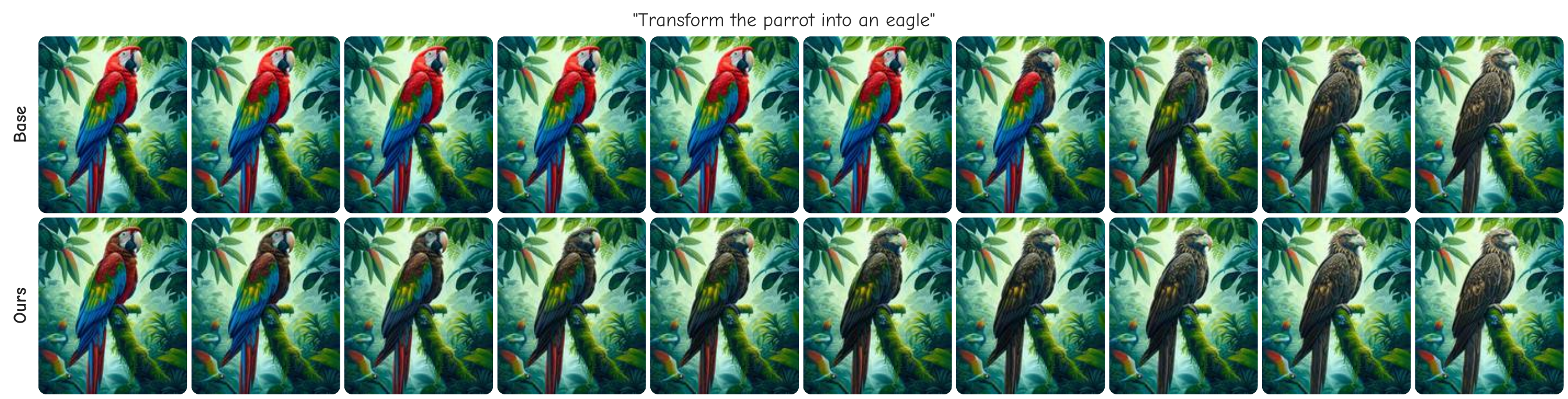}
    \includegraphics[width=0.95\textwidth]{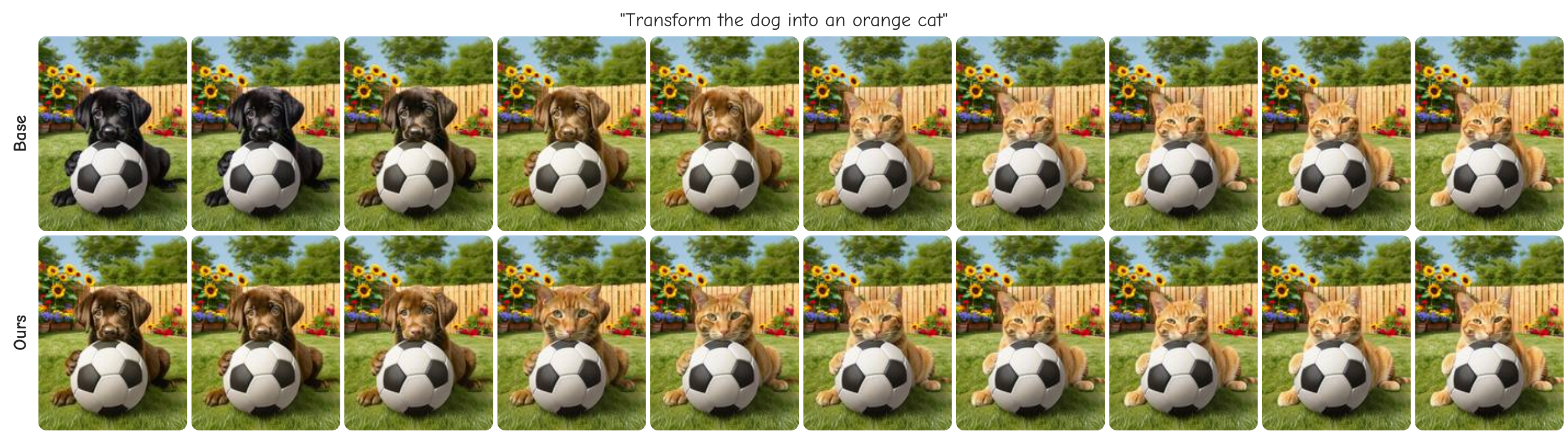}
    \includegraphics[width=0.95\textwidth]{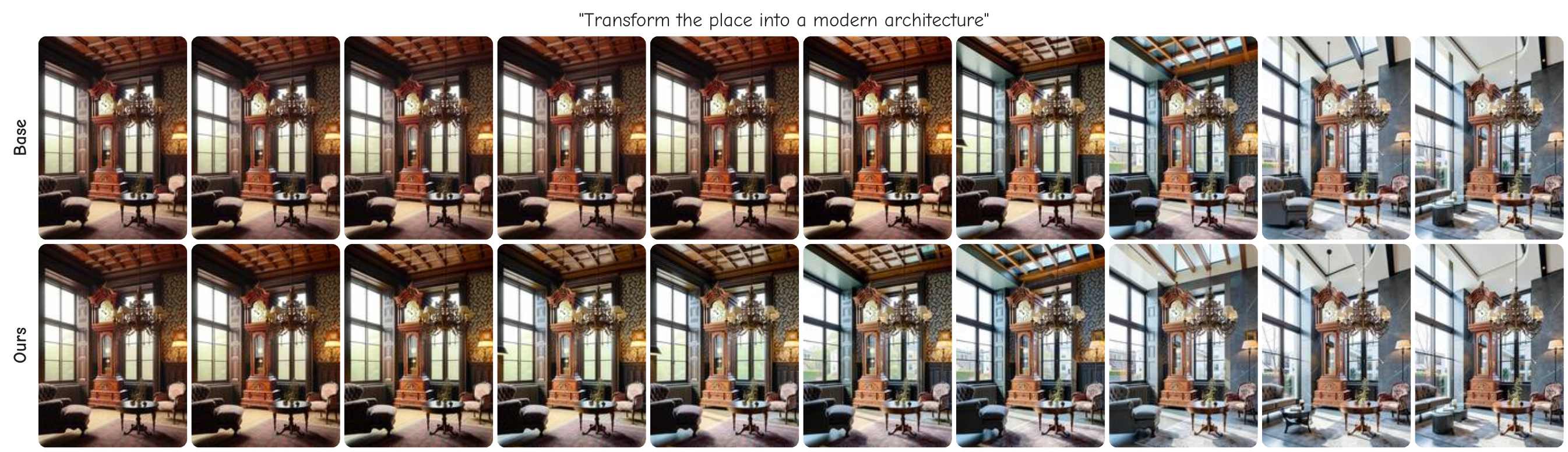}
    \includegraphics[width=0.95\textwidth]{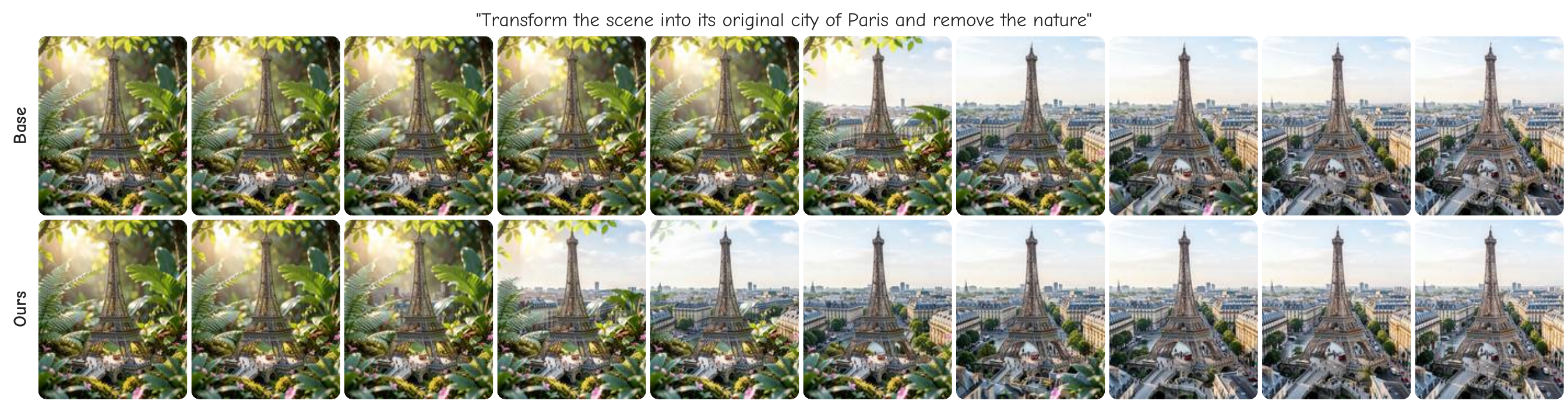}
    \includegraphics[width=0.95\textwidth]{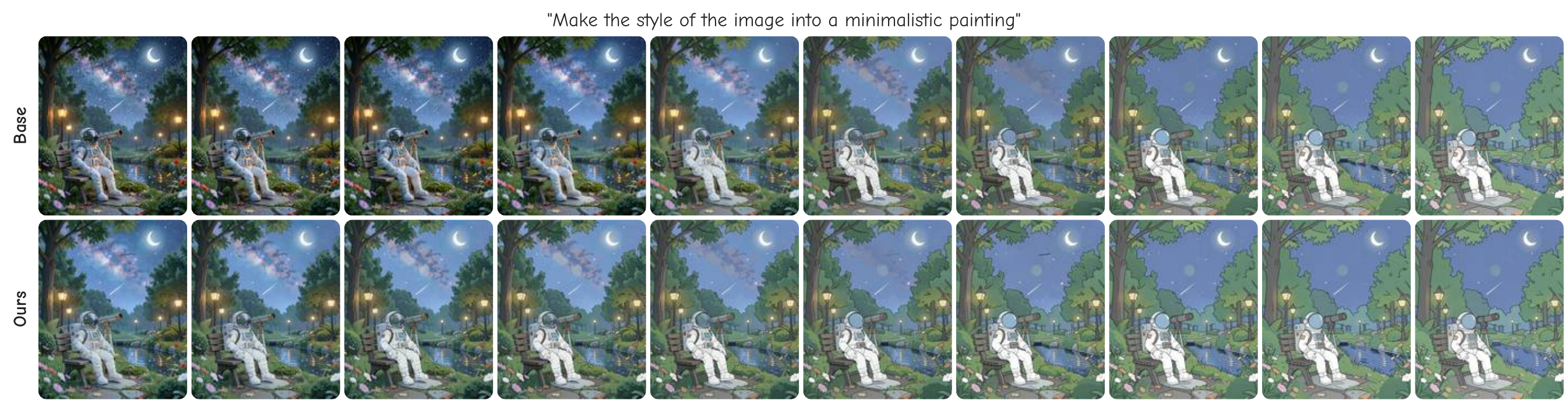}
    \includegraphics[width=0.95\textwidth]{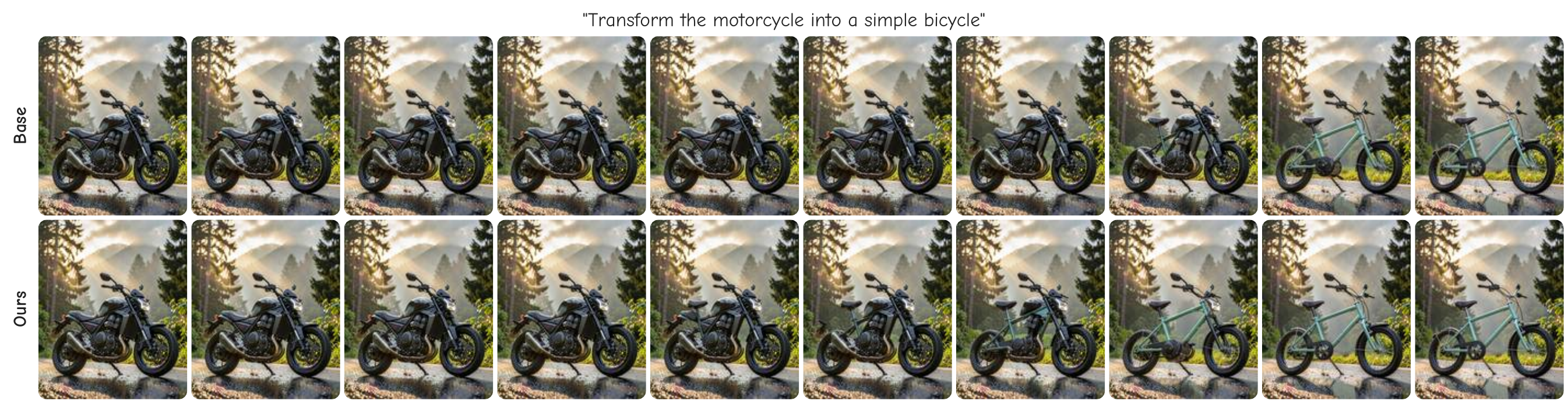}
    \caption{Qualitative comparison demonstrating the effect of our regularizer relative to the baseline SliderEdit method under the same backbone.}
    \label{fig:qualitative_examples_with_and_without_regularizers_appendix_base}
\end{figure}

\begin{figure}[t]
    \centering
    \includegraphics[width=0.9\textwidth]{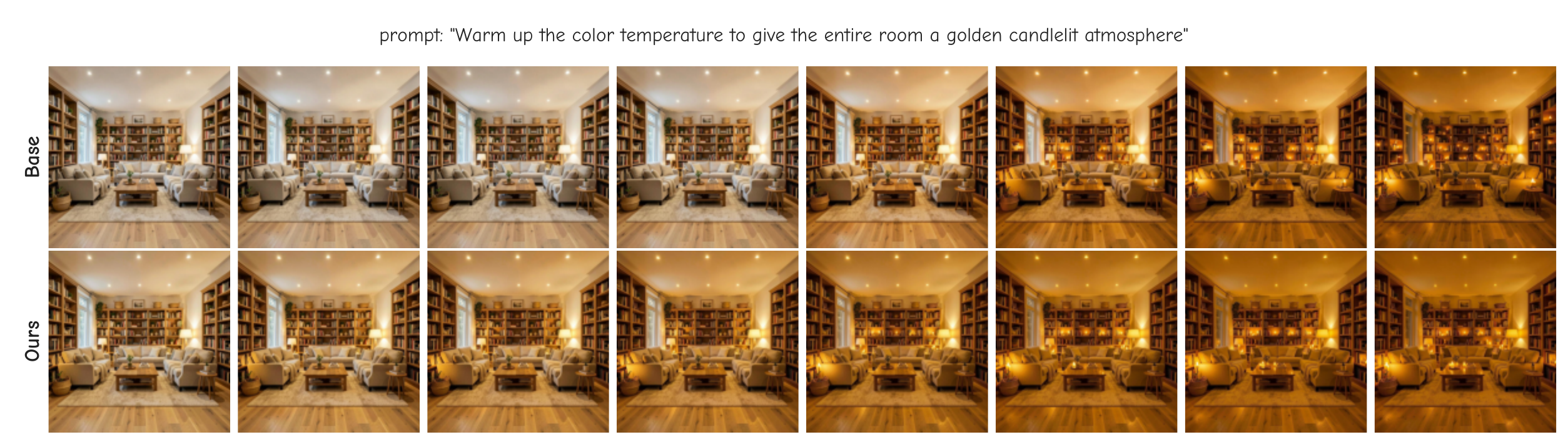}
    \includegraphics[width=0.9\textwidth]{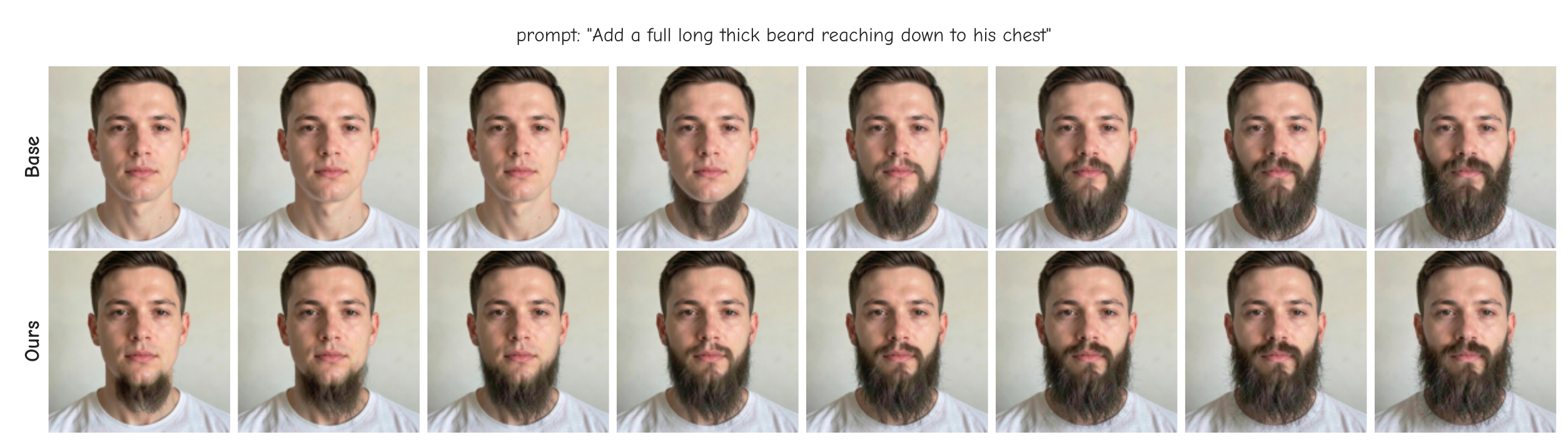}
    \includegraphics[width=0.9\textwidth]{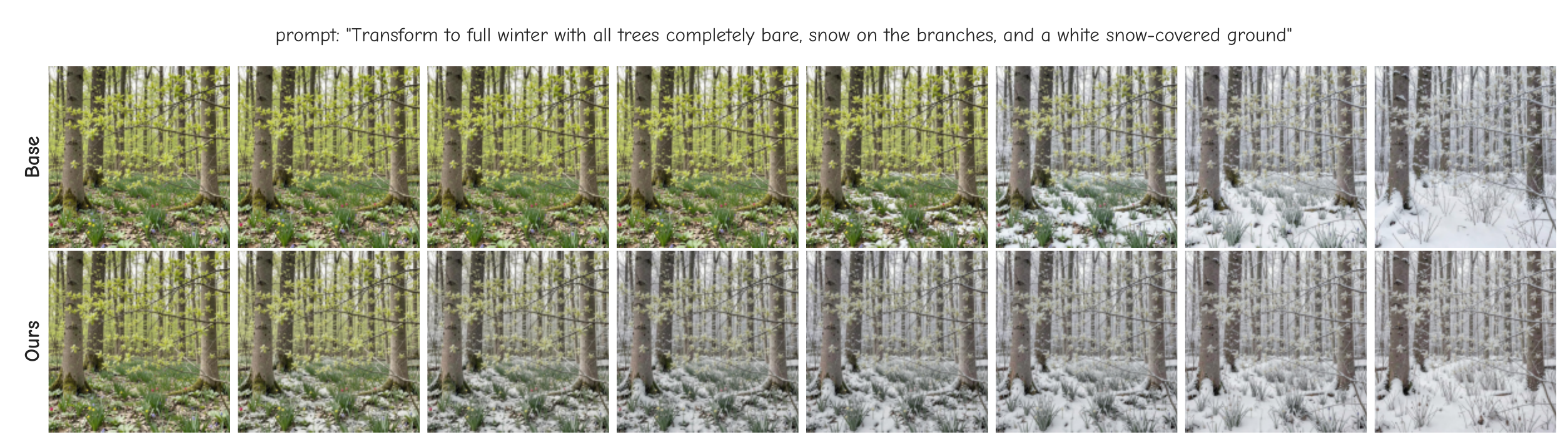}
    \includegraphics[width=0.9\textwidth]{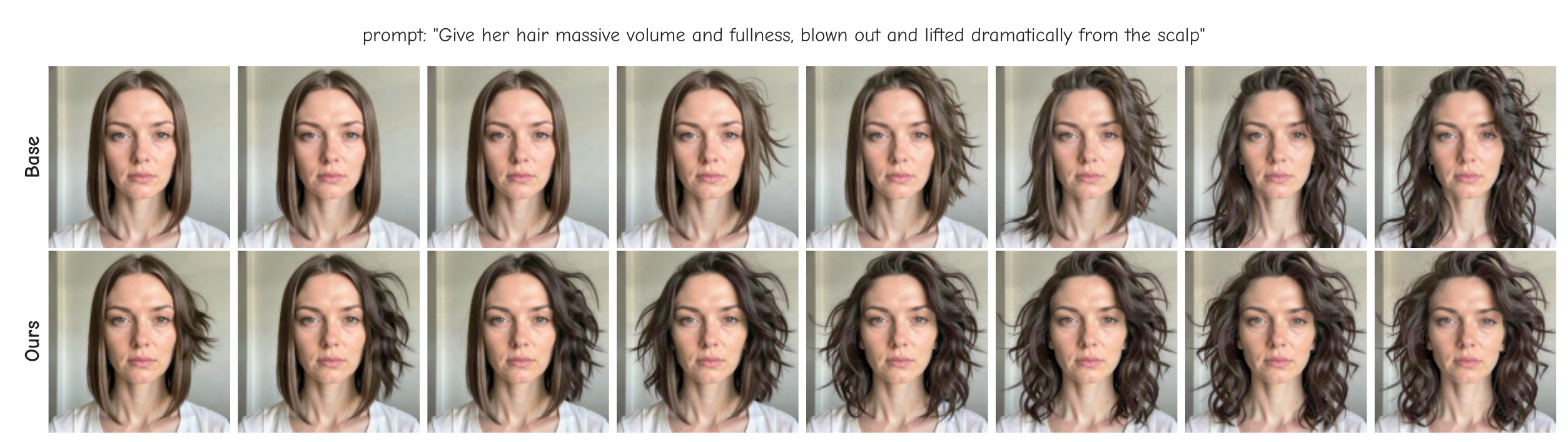}
    \includegraphics[width=0.9\textwidth]{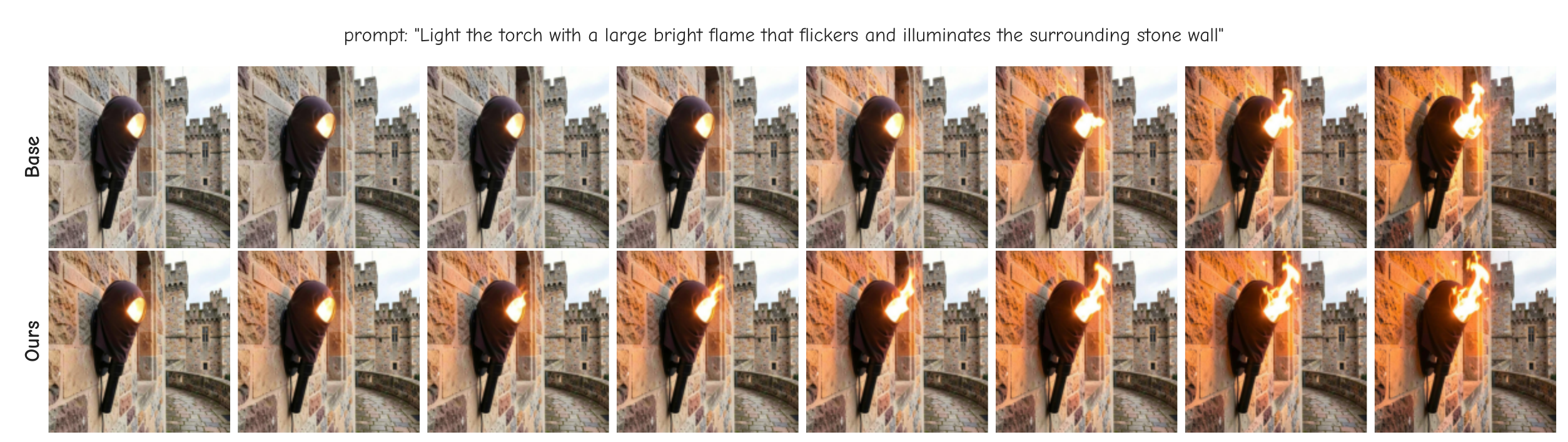}
    \includegraphics[width=0.9\textwidth]{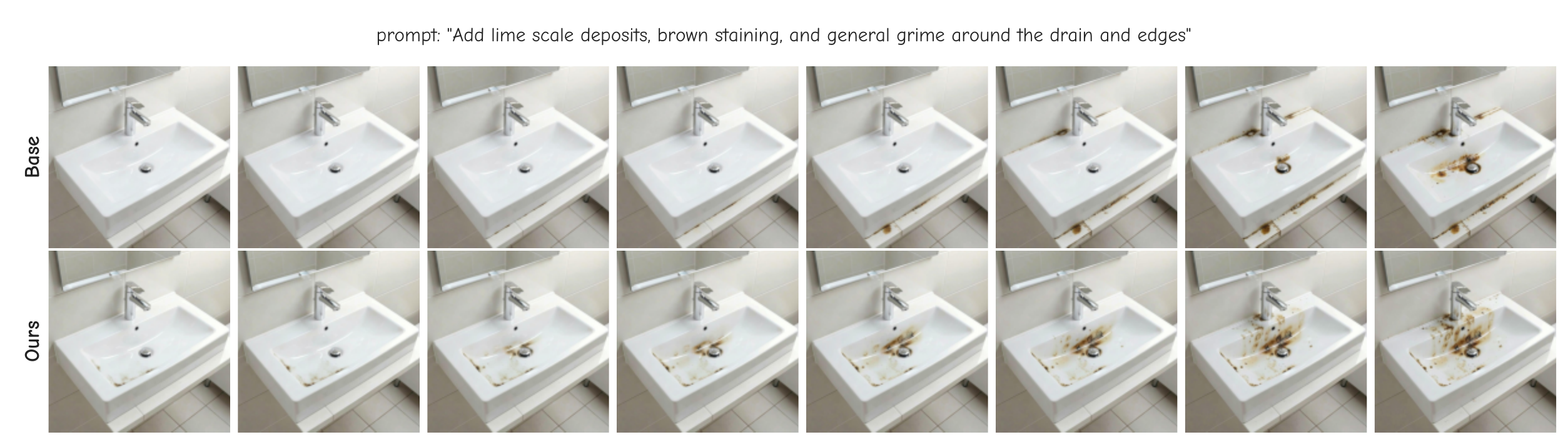}
    \caption{Qualitative comparison demonstrating the effect of our regularizer relative to the baseline SliderEdit method under the same backbone.}
    \label{fig:qualitative_examples_with_and_without_regularizers_appendix_base_2}
\end{figure}

\begin{figure}[t]
    \centering
    \includegraphics[width=\textwidth]{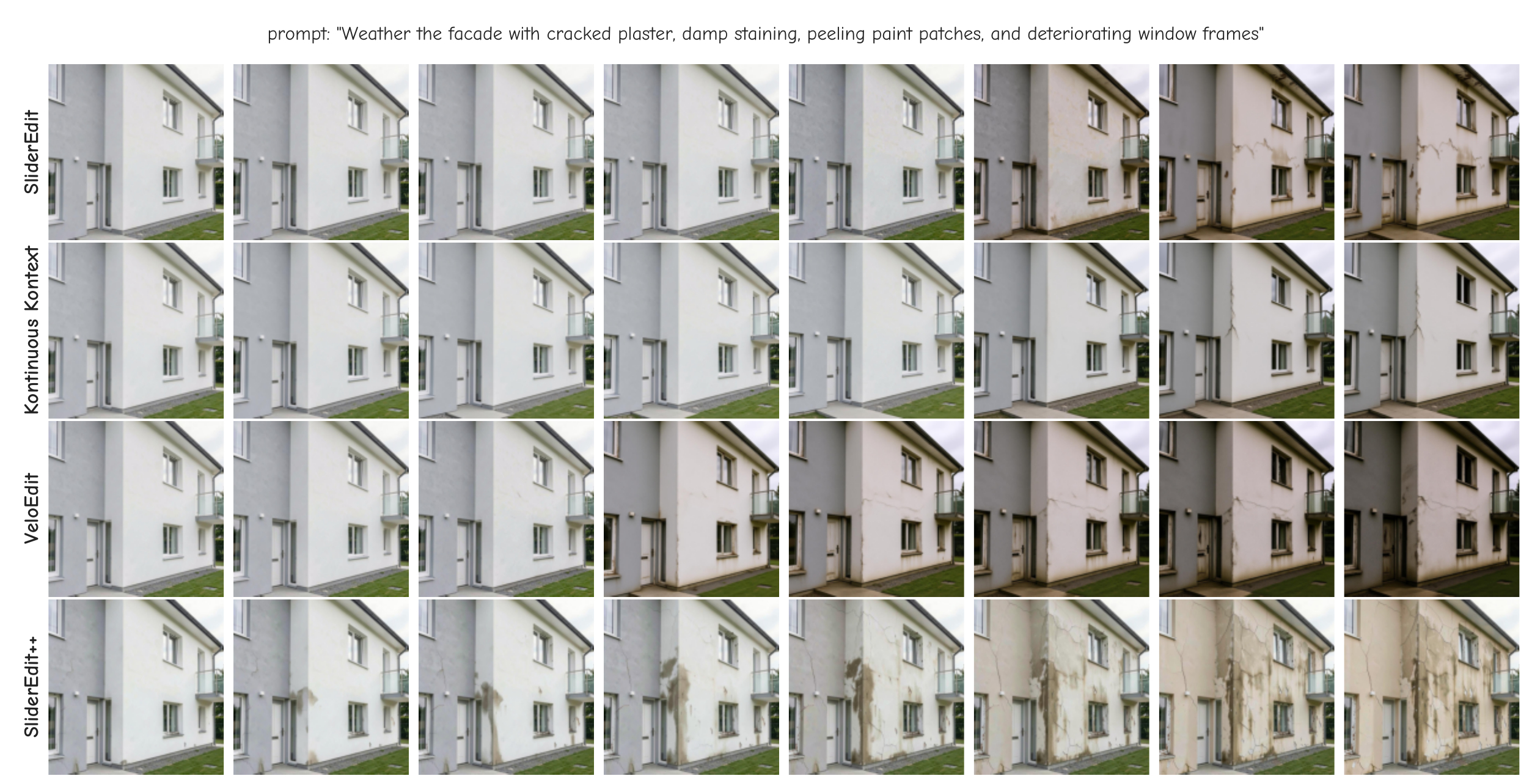}
    \includegraphics[width=\textwidth]{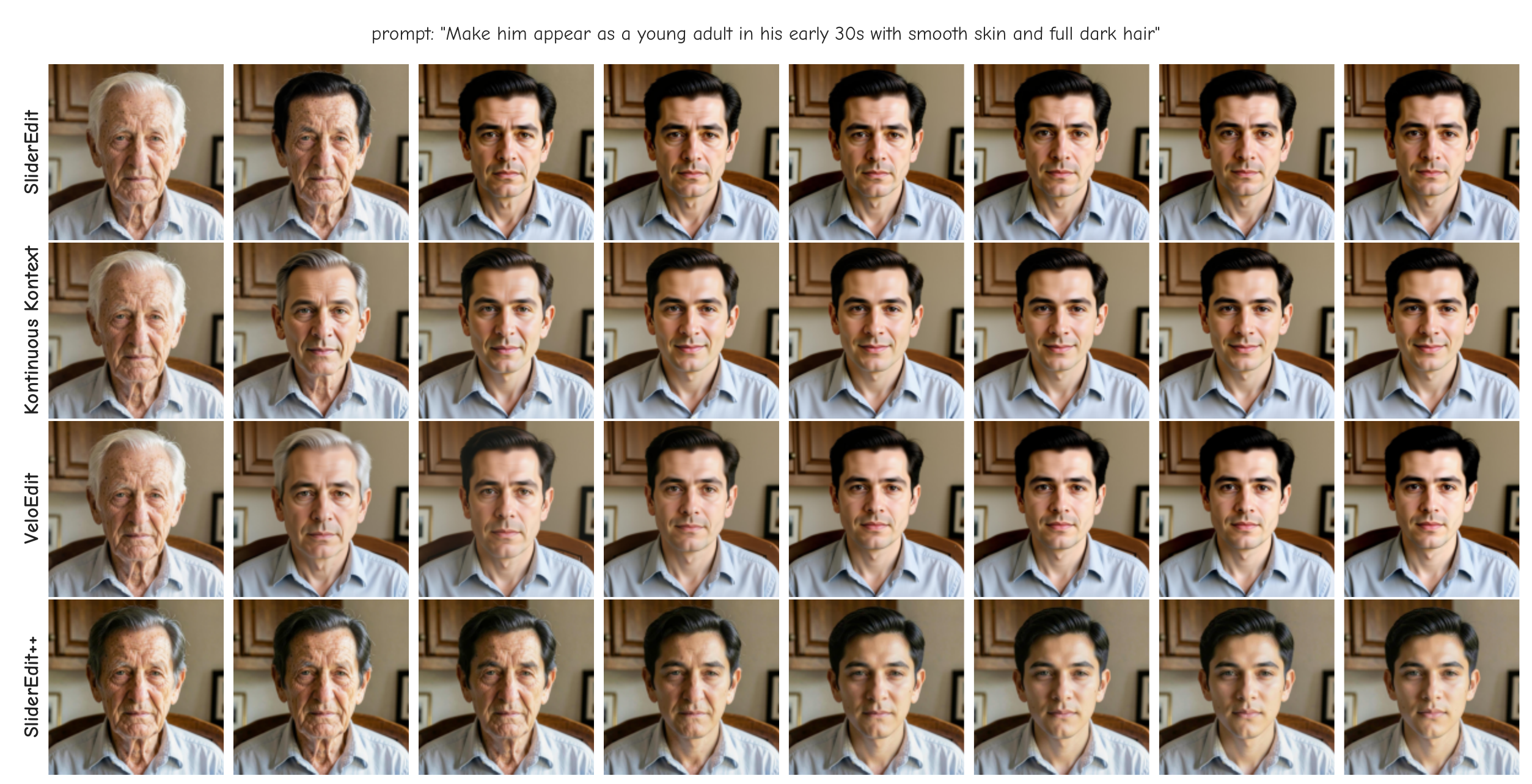}
    \includegraphics[width=\textwidth]{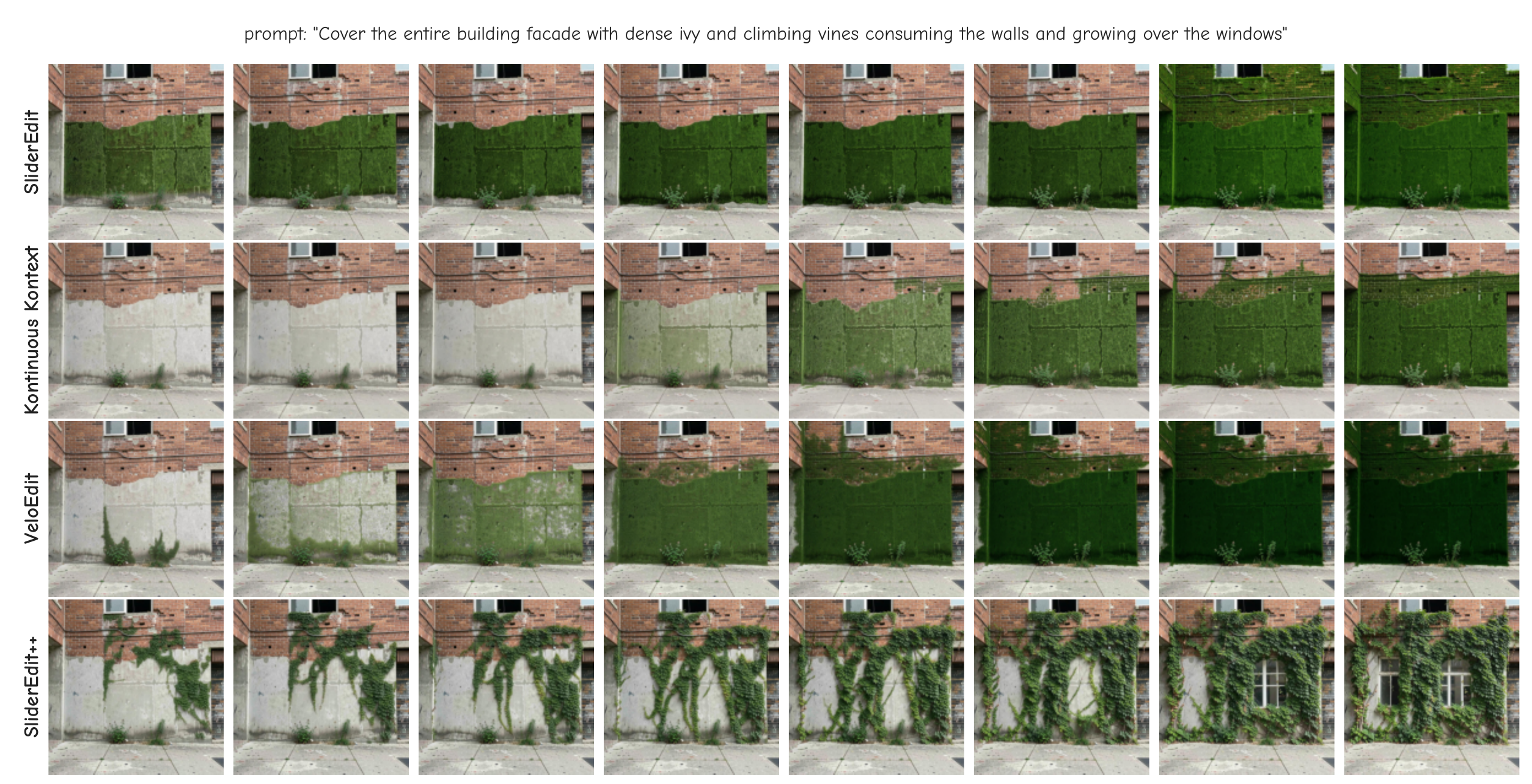}
    \caption{Qualitative comparison of continuous edit-control methods. \modelname\ demonstrates superior performance.}
    \label{fig:qualitative_examples_comparison_baselines_appendix}
\end{figure}

\begin{figure}[t]
    \centering
    \includegraphics[width=0.49\linewidth]{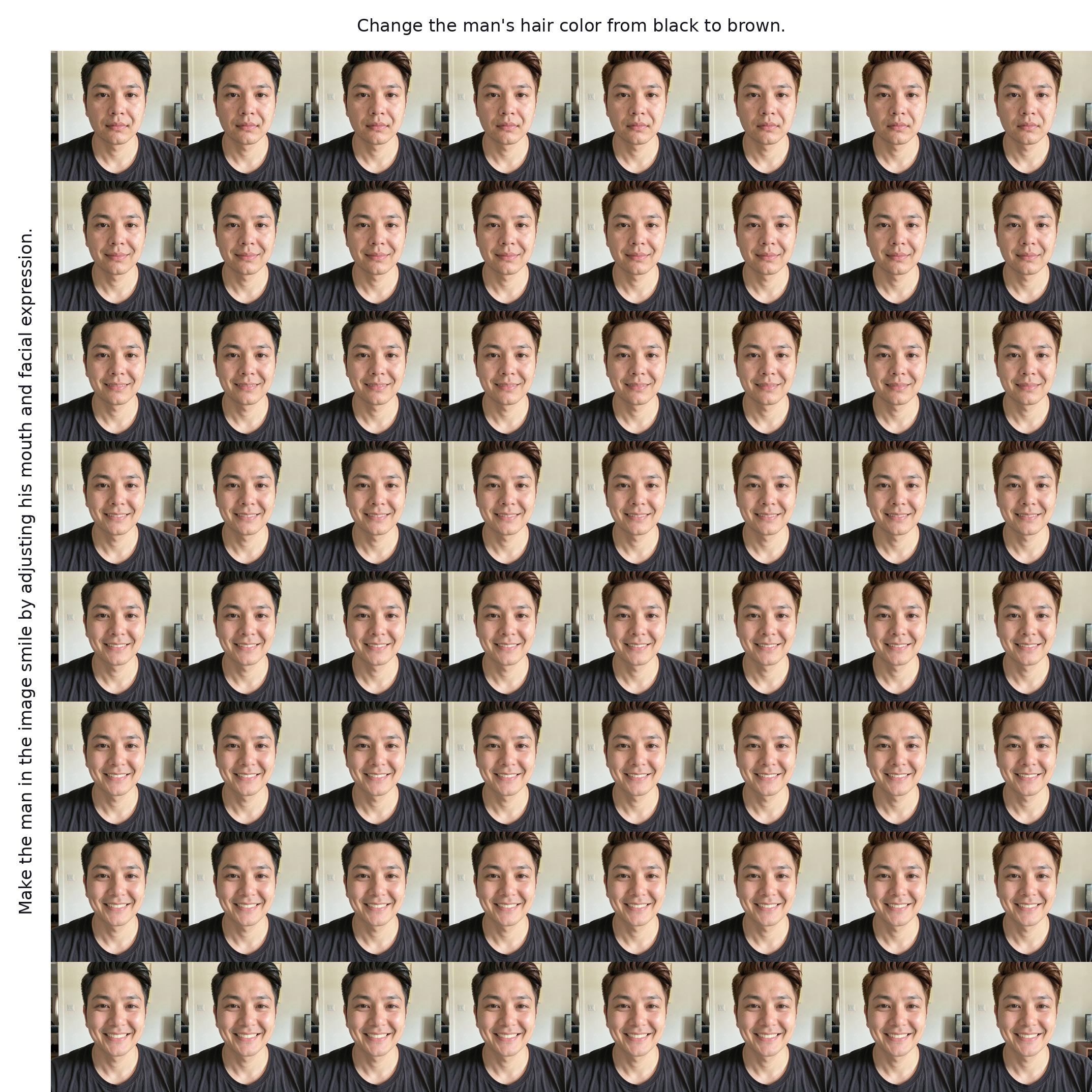}
    \hfill
    \includegraphics[width=0.49\linewidth]{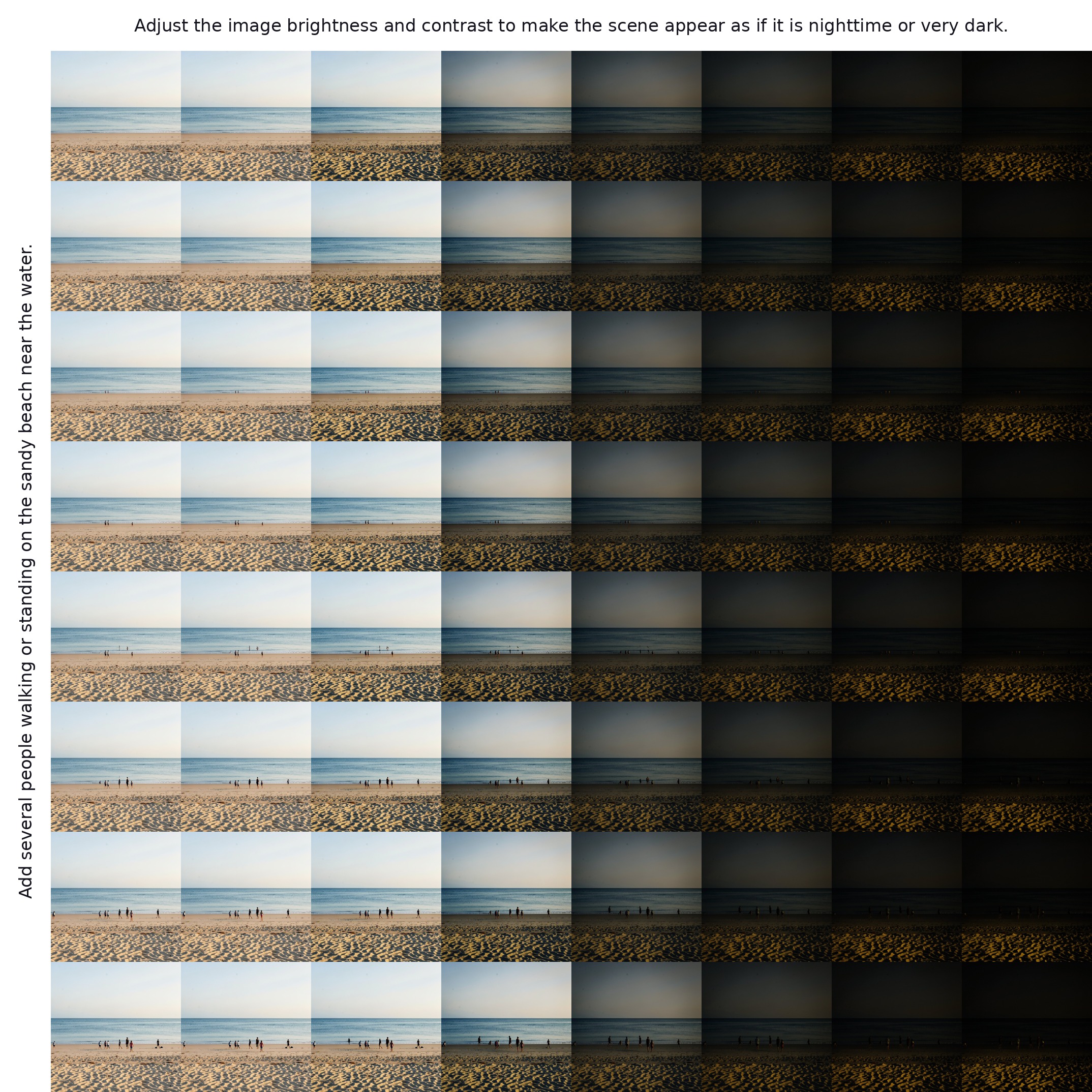}

    \vspace{2pt}

    \includegraphics[width=0.49\linewidth]{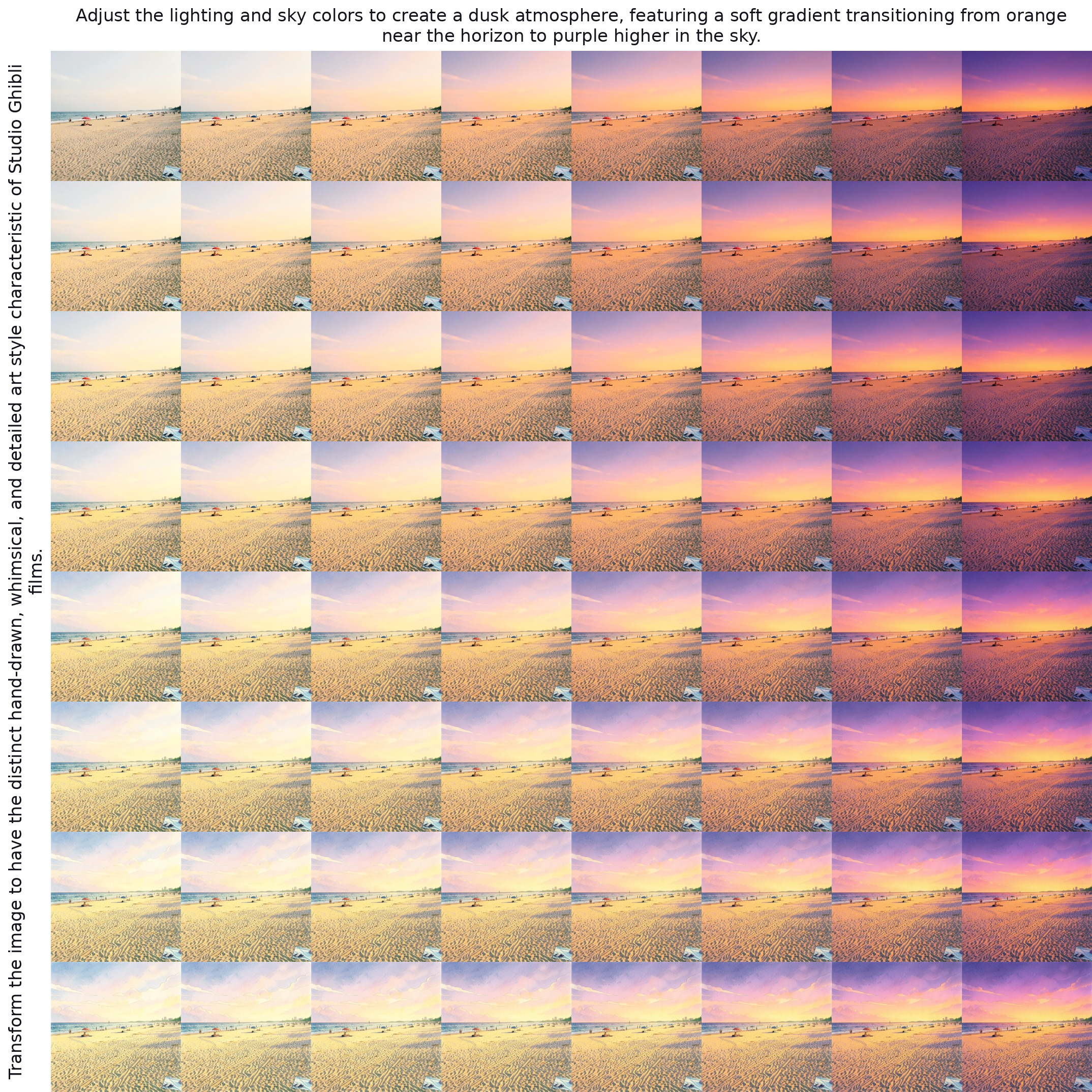}
    \hfill
    \includegraphics[width=0.49\linewidth]{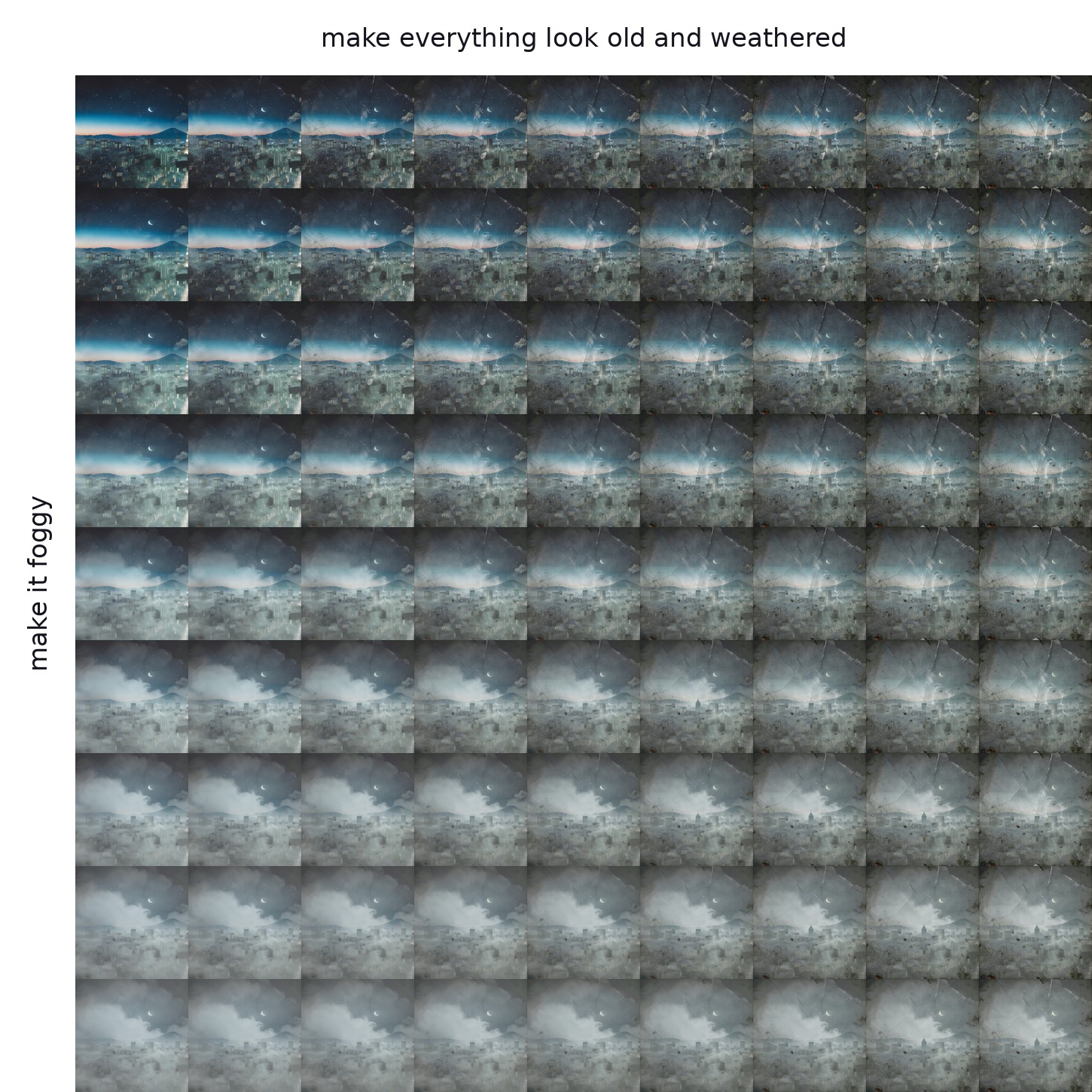}
    \caption{Qualitative Results on Multi-Slider Outputs. Best viewed zoomed in.}
    \label{fig:multi_slider_qualitative_appendix}
\end{figure}

\begin{figure}
    \centering
    \includegraphics[width=\linewidth]{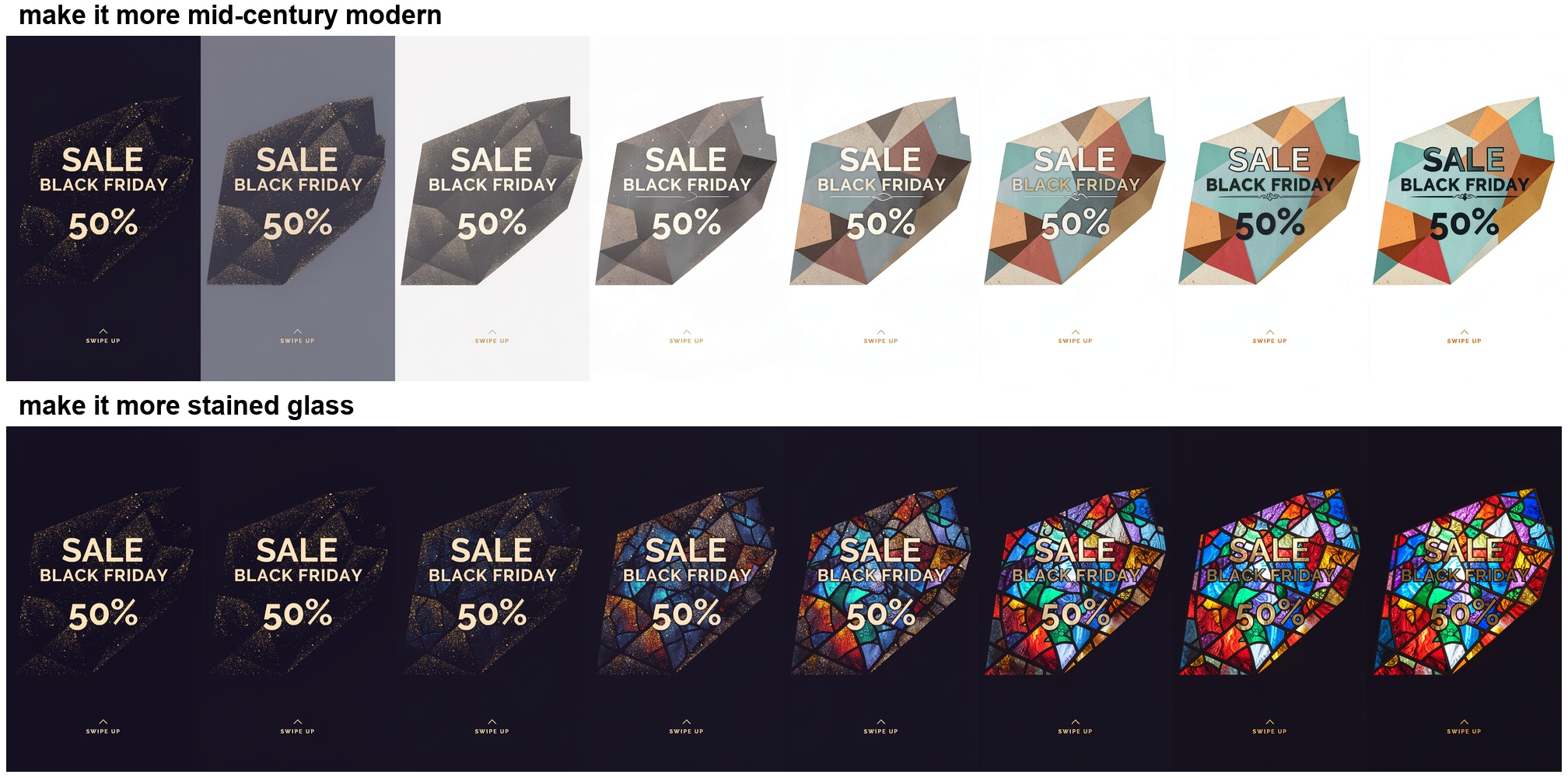}
    \caption{Qualitative Results on Design Images. }
    \label{fig:design_qualitative_appendix}
  \end{figure}

\begin{figure}
\centering
\includegraphics[width=\linewidth]{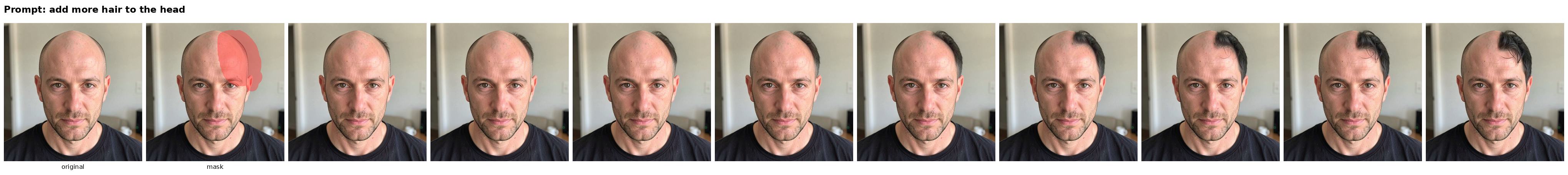}
\includegraphics[width=\linewidth]{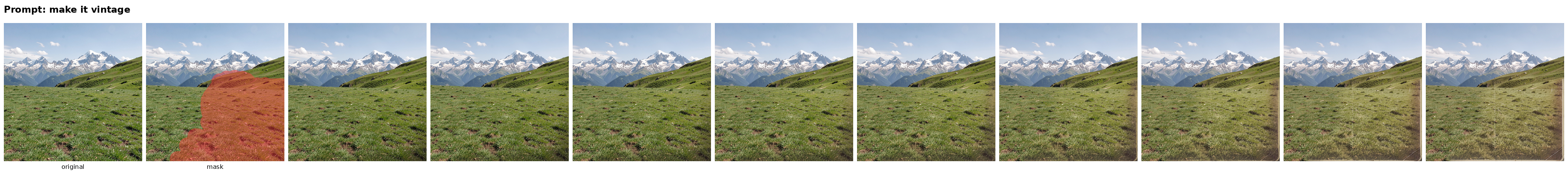}
\includegraphics[width=\linewidth]{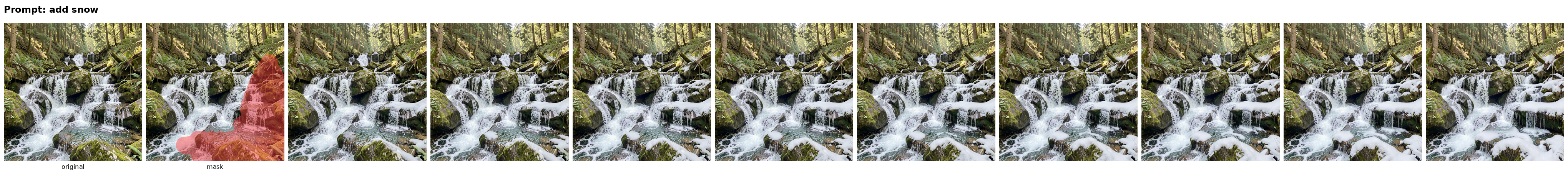}
\includegraphics[width=\linewidth]{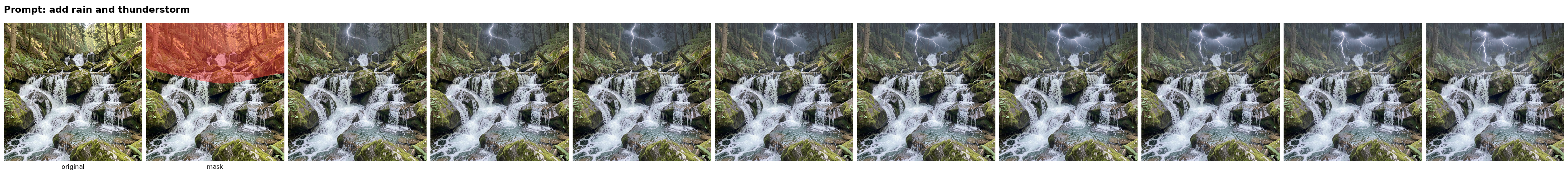}
\includegraphics[width=\linewidth]{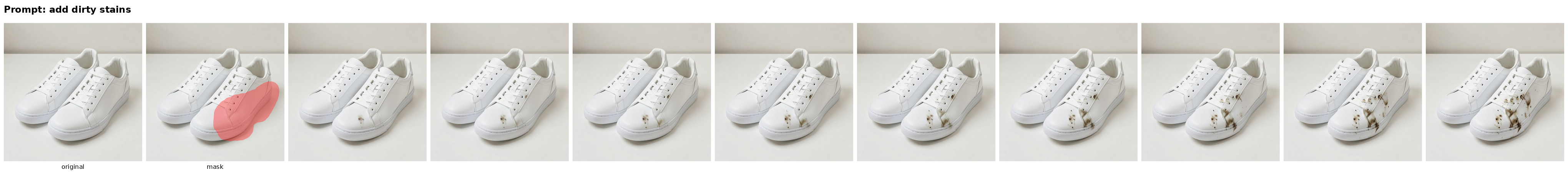}
\includegraphics[width=\linewidth]{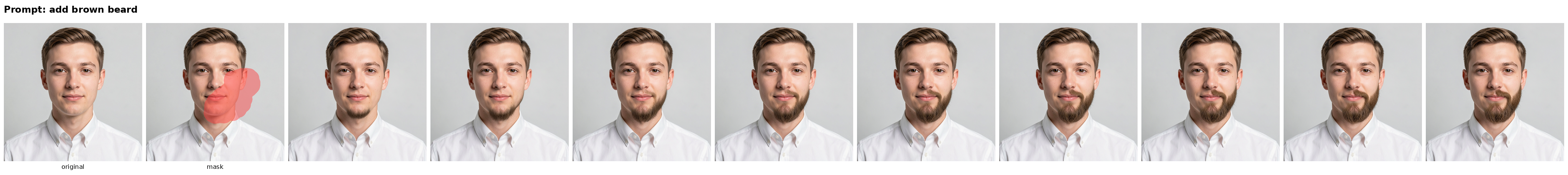}
\includegraphics[width=\linewidth]{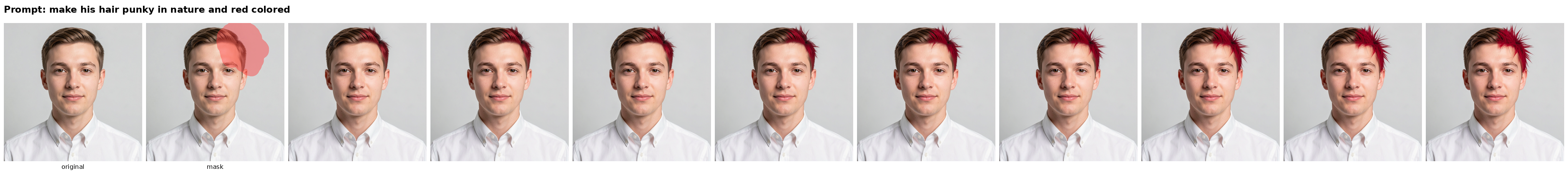}
\vspace{-12pt}
\caption{Qualitative Results on Mask-Conditioned \modelname{}.}
\label{fig:masked_conditioned_qualitative_appendix}
\end{figure}

\vspace{-4pt}
Figure~\ref{fig:qualitative_examples_appendix} presents additional qualitative examples of \modelname{}, demonstrating smooth and continuous single-slider control over both global edits and fine-grained local changes. Figures~\ref{fig:qualitative_examples_with_and_without_regularizers_appendix_base} and~\ref{fig:qualitative_examples_with_and_without_regularizers_appendix_base_2} illustrate the effect of our proposed regularizers compared with the base training objective without regularization. While the base objective often exhibits abrupt transitions between neighboring slider values, our regularizers yield substantially more semantically continuous editing trajectories. Figure~\ref{fig:qualitative_examples_comparison_baselines_appendix} provides additional comparisons with existing baselines, where \modelname{} consistently produces smoother and more continuous control. Figure~\ref{fig:multi_slider_qualitative_appendix} presents multi-slider examples, showing that the learned controls remain largely disentangled while maintaining smooth transitions along each editing dimension. Figure~\ref{fig:design_qualitative_appendix} demonstrates the applicability of \modelname{} to text-rich design posters, where sliders provide continuous control over visual themes while preserving the underlying content and layout. Finally, Figure~\ref{fig:masked_conditioned_qualitative_appendix} shows additional examples of our mask-conditioned extension, enabling spatially localized continuous editing within a user-selected region.

\vspace{-4pt}
\subsection{Ablating on few-step generation}
\vspace{-4pt}
We ablate the number of denoising steps used by our few-step model. While our default setting uses $N=8$ steps, we further reduce it to $N=4$, $2$, and $1$. As shown in Table~\ref{tab:ablating_denoising_steps}, \modelname{} maintains strong performance even with substantially fewer steps. This suggests that the inference latency can be further reduced by up to $8\times$, enabling an even more responsive interactive editing experience.

\begin{figure}
\vspace{-8pt}
\centering
\includegraphics[width=0.5\linewidth]{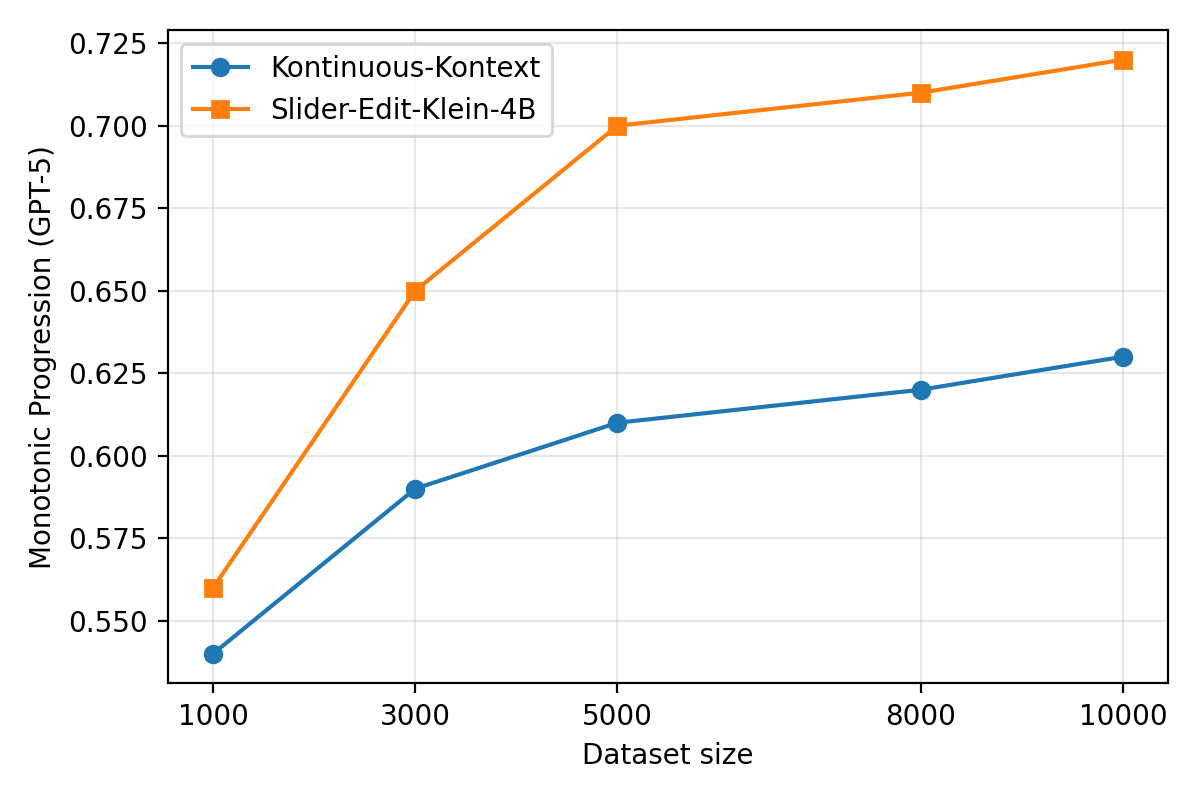}
\vspace{-12pt}
\caption{At similar data-budgets, SliderEdit on a stronger base model such as Klein-4B outperforms Kontinuous-Kontext on the same Klein-4B model.}
\label{fig:data_budget_experiment}
\end{figure}

\vspace{-4pt}
\subsection{Ablating the Underlying Slider Training Objective}
\vspace{-4pt}

Our regularizers can be applied on top of different slider training objectives. We compare the two primary candidates, SliderEdit and Kontinuous-Kontext, which differ substantially in their data requirements: Kontinuous-Kontext uses approximately 110k paired samples, while SliderEdit requires $\leq$10k unpaired samples. As shown in Fig.~\ref{fig:data_budget_experiment}, on the few-step Klein-4B backbone, SliderEdit achieves stronger performance than Kontinuous-Kontext at comparable data budgets. We therefore adopt SliderEdit as our base objective, as it provides stronger performance while avoiding the need for paired training data.

\vspace{-4pt}
\section{Limitations}
\vspace{-4pt}
\label{sec:app:limitations}
While \modelname{} consistently improves continuous editing across categories, its performance remains constrained by the capabilities of the underlying image-editing model. In particular, challenging categories such as \emph{Camera} achieve lower absolute performance across all methods, suggesting that some limitations stem from the base model's ability to perform the corresponding edits rather than from slider control itself. Improving the underlying editing model could therefore further enhance \modelname{} in these challenging categories.


\end{document}